%% file: arxiv.tex
\documentclass{article} 
\usepackage{arxiv,times}

\input{math_commands.tex}

\usepackage[T1]{fontenc}

\usepackage[utf8]{inputenc}

\usepackage{microtype}

\usepackage{inconsolata}

\usepackage{algorithm}
\usepackage{algpseudocode}
\usepackage{amsmath,amssymb,amsthm}
\usepackage{bm}
\usepackage{booktabs}
\usepackage{colortbl}
\usepackage{dsfont}
\usepackage{enumitem}
\usepackage{fancyvrb}
\usepackage{float}
\usepackage[T1]{fontenc}
\usepackage{geometry}
\usepackage{graphicx}
\usepackage{hyperref}
\usepackage[utf8]{inputenc}
\usepackage{lineno}
\usepackage{listings}
\usepackage{makecell}
\usepackage{microtype}
\usepackage{multirow}
\usepackage{multicol}
\usepackage{pgfmath}
\usepackage{pifont}
\usepackage{tabularx}
\usepackage{tcolorbox}
\usepackage{url}
\usepackage{wrapfig}
\usepackage{xcolor}
\usepackage{xspace}

\title{Pay for the Fault, Not the Flow: Label-Free In-Flow Multi-Agent Workflow Optimization}

\author{
\makebox[\dimexpr\textwidth-2\tabcolsep\relax][c]{Xuehang Guo\textsuperscript{1}, Haoyu Wang\textsuperscript{2}, Shengyu Chen\textsuperscript{2}, Zach Chen\textsuperscript{2}, Wei Cheng\textsuperscript{2}, Qingyun Wang\textsuperscript{1}, Haifeng Chen\textsuperscript{2}} \\
\makebox[\dimexpr\textwidth-2\tabcolsep\relax][c]{\textsuperscript{1}William \& Mary \quad \textsuperscript{2}NEC Corporation of America}
}

\input{sections/macro}

\renewenvironment{abstract}
  {\begin{tcolorbox}[colback=bboxface,colframe=bboxface,boxrule=0pt,arc=6pt,
                     left=14pt,right=14pt,top=10pt,bottom=10pt,boxsep=0pt,
                     width=\textwidth,grow to left by=24pt,grow to right by=24pt,
                     before skip=-0.1in,after skip=\medskipamount]
   \centerline{\large\sc Abstract}\vspace{0.6ex}}
  {\end{tcolorbox}\vskip 1ex}

\iclrfinalcopy 
\begin{document}

\vspace*{-0.45in}
\maketitle

\makeatletter
\begingroup
  \renewcommand{\thefootnote}{}%
  \long\def\@makefntext#1{\parindent 0pt\noindent#1}%
  \footnotetext{Xuehang Guo \texttt{<xguo15@wm.edu>}, Haoyu Wang \texttt{<haoyu@nec-labs.com>}, Haifeng Chen \texttt{<haifeng@nec-labs.com>}}%
\endgroup
\makeatother

\begin{abstract}
Large language models (LLMs) increasingly construct multi-agent workflows that decompose a complex task and assign specialist agents from a pool. However, building such a workflow well remains challenging: how finely to divide the task, which agent to trust with each subtask, and when to create a new specialist are all critical decisions a workflow constructor needs to settle up front. Thus, whether each subtask succeeds remains unknown until the workflow runs. Yet, improving a workflow is costly. Locating a fault usually requires a reference answer, a graded outcome, or a trained assessor, and the fix is applied to the whole workflow through re-execution, re-search, or retraining. We propose \ours, which prices every decision in one label-free cost that weighs how well an agent's competence meets what a subtask demands against how much that agent takes to run. Before execution, \ours bidirectionally determines the granularity of task decomposition and agent assignment following from the cost rather than from a fixed template. During execution, \ours corrects a fault with the cheapest move via the same cost that serves the workflow both as it is built and as it runs. Facing the workflow-level evaluation challenge, we introduce \braid, a benchmark whose tasks require multi-agent coordination beyond single-agent capability. Across various domains and backbones, \ours outperforms single agent baselines by up to $+11.97\%$, achieving $+9.64\%$ with in-flow optimization. Our project page: {\hypersetup{urlcolor=oursttcolor}\url{https://xhguo7.github.io/InFlowOp/}}.
\end{abstract}

\input{sections/1_intro}
\input{sections/2_prelim}
\input{sections/3_method}
\input{sections/4_braid}
\input{sections/5_exps}
\input{sections/8_end}

\clearpage


\bibliography{iclr2027_conference}
\bibliographystyle{iclr2027_conference}

\input{sections/appendix}

\end{document}

%% file: math_commands.tex
\usepackage{amsmath,amsfonts,bm}

\def\eqref#1{equation~\ref{#1}}

\def\1{\bm{1}}

\DeclareMathAlphabet{\mathsfit}{\encodingdefault}{\sfdefault}{m}{sl}
\SetMathAlphabet{\mathsfit}{bold}{\encodingdefault}{\sfdefault}{bx}{n}



%% file: sections/macro.tex
\definecolor{ourscolor}{HTML}{192B36}
\definecolor{oursttcolor}{HTML}{1C5EC2}

\definecolor{commentcolor}{HTML}{5C85B4}

\definecolor{mycitecolor}{HTML}{CA78D6}

\definecolor{bboxedge}{HTML}{96B3D6}
\definecolor{bboxface}{HTML}{F2F7FB}
\definecolor{bboxtag}{HTML}{80A3CE}

\definecolor{notunecolor}{HTML}{888888}

\definecolor{sandboxcolor}{HTML}{EF8537}
\definecolor{teachergreen}{HTML}{5CB86F}

\definecolor{gray_bg}{HTML}{F2F2F2}
\definecolor{gray_text}{HTML}{7D7D7D}

\definecolor{blue_bg}{HTML}{EAF1F9}
\definecolor{blue_text}{HTML}{4F80BE}
\definecolor{blue_delta}{HTML}{4F80BE}

\definecolor{teal_bg}{HTML}{E7F2EF}
\definecolor{teal_text}{HTML}{4F9484}
\definecolor{teal_delta}{HTML}{4F9484}

\definecolor{green_bg}{HTML}{EBF3E7}
\definecolor{green_text}{HTML}{5C9E44}
\definecolor{green_delta}{HTML}{5C9E44}

\definecolor{purple_bg}{HTML}{F1EAF7}
\definecolor{purple_text}{HTML}{8E6EAA}
\definecolor{purple_delta}{HTML}{8E6EAA}

\definecolor{orange_bg}{HTML}{FBF0E4}
\definecolor{orange_text}{HTML}{D2792A}
\definecolor{orange_delta}{HTML}{D2792A}

\definecolor{baseline_bg}{HTML}{E6E6E6}

\newcommand{\gaintint}[2]{%
  \pgfmathsetmacro{\gainres}{#2-#1}%
  \pgfmathtruncatemacro{\gainint}{int(max(6, min(34, 8 + abs(\gainres))))}%
  \ifdim\gainres pt>0pt\edef\gaincolor{blue_delta!\gainint!white}%
  \else\edef\gaincolor{white}\fi
  \expandafter\cellcolor\expandafter{\gaincolor}%
}
\newcommand{\gainfmt}[1]{%
  \pgfmathtruncatemacro{\gainwhole}{int(#1)}%
  \pgfmathtruncatemacro{\gaindec}{int(round((abs(#1)-abs(\gainwhole))*100))}%
  \ifnum\gaindec<10\edef\gaindecstr{0\gaindec}\else\edef\gaindecstr{\gaindec}\fi%
  \gainwhole.\gaindecstr%
}
\newcommand{\gaincell}[2]{\gaintint{#1}{#2}#2}
\newcommand{\gaindeltaflat}[2]{%
  \pgfmathsetmacro{\gainres}{#2-#1}%
  \textbf{\textcolor{blue_text}{\ifdim\gainres pt>0pt$+$\fi\gainfmt{\gainres}}}%
}
\newcommand{\gaindelta}[2]{%
  \pgfmathsetmacro{\gainres}{#2-#1}%
  \gaintint{#1}{#2}\textbf{\textcolor{blue_text}{\ifdim\gainres pt>0pt$+$\fi\gainfmt{\gainres}}}%
}

\definecolor{deltagreen}{HTML}{80C577}
\definecolor{deltared}{HTML}{EF8787}
\definecolor{deltaredbg}{HTML}{F4AAAA}
\colorlet{deltacellcolor}{white} 

\definecolor{checkgreen}{HTML}{31B331}
\definecolor{crossred}{HTML}{E6716E}
\definecolor{checkgreen_bg}{HTML}{DFF6DF}
\definecolor{crossred_bg}{HTML}{FDE6E6}

\hypersetup{
    colorlinks=true,
    linkcolor=mycitecolor,
    citecolor=mycitecolor,
    urlcolor=mycitecolor
}

\newcommand{\ours}{\textcolor{ourscolor}{\textsc{InFlowOp}}\xspace}

\newcommand{\coalesce}{\textcolor{ourscolor}{\textsc{Coalesce}}\xspace}

\newcommand{\braid}{\textcolor{ourscolor}{\textsc{Braid}}\xspace}

\tcbuselibrary{skins,breakable}

\newtcolorbox{hlbox}[1]{
  enhanced, breakable,
  colback=bboxface, colframe=bboxface, boxrule=0pt, arc=6pt,
  left=8pt, right=8pt, top=12pt, bottom=6pt,
  title=#1,
  fonttitle=\footnotesize\bfseries, coltitle=white,
  attach boxed title to top left={xshift=10pt, yshift=-\tcboxedtitleheight/2},
  boxed title style={colback=bboxtag, colframe=bboxtag, boxrule=0pt, arc=3pt,
                     left=6pt, right=6pt, top=2pt, bottom=2pt},
}

\newcommand{\circnum}[1]{%
  \begin{tikzpicture}[baseline=-0.6ex]
    \node[circle, fill=bboxedge, inner sep=0.6pt,
          minimum size=6pt, text=white,
          font=\bfseries\small] {#1};
  \end{tikzpicture}%
}

\newlist{oursenumerate}{enumerate}{1}

\setlist[oursenumerate,1]{
  label=\protect\circnum{\arabic*},
  leftmargin=1.5em,
  itemsep=0pt,
  topsep=0pt,
  partopsep=0pt,
  parsep=0pt,
}

\algrenewcommand{\algorithmiccomment}[1]{\hfill\textcolor{gray}{\(\triangleright\) #1}}

\newcommand{\one}{\mathds{1}}

\newcommand{\applydeltacellcolor}[1]{%
  \pgfmathsetmacro{\myintensity}{int(min(100, abs(#1) * 6))}%
  \ifdim#1pt>0pt%
    \edef\tmpcolor{deltagreen!\myintensity!white}%
  \else%
    \edef\tmpcolor{deltaredbg!\myintensity!white}%
  \fi%
  \expandafter\cellcolor\expandafter{\tmpcolor}%
}

\newcommand{\myul}[1]{\leavevmode\vtop{\vbox{\hbox{#1}}\hrule height 0.4pt}}

\newcommand{\testvaldeltabgpct}[2]{%
  \pgfmathsetmacro{\deltares}{#2-#1}%
  \pgfmathtruncatemacro{\valwhole}{int(#2)}%
  \pgfmathtruncatemacro{\valdec}{int(round((#2-\valwhole)*100))}%
  \ifnum\valdec<10\edef\valdecstr{0\valdec}\else\edef\valdecstr{\valdec}\fi%
  \ifdim\deltares pt>0pt%
    \pgfmathsetmacro{\deltaint}{int(min(100, abs(\deltares) * 3))}%
    \edef\deltacolor{deltagreen!\deltaint!white}%
    \expandafter\cellcolor\expandafter{\deltacolor}\small\rmfamily\valwhole.\valdecstr%
  \else%
    \pgfmathsetmacro{\deltaint}{int(min(100, abs(\deltares) * 2))}%
    \edef\deltacolor{deltared!\deltaint!white}%
    \expandafter\cellcolor\expandafter{\deltacolor}\small\rmfamily\valwhole.\valdecstr%
  \fi%
}

\newcommand{\testvaldeltabgpctbf}[2]{%
  \pgfmathsetmacro{\deltares}{#2-#1}%
  \pgfmathtruncatemacro{\valwhole}{int(#2)}%
  \pgfmathtruncatemacro{\valdec}{int(round((#2-\valwhole)*100))}%
  \ifnum\valdec<10\edef\valdecstr{0\valdec}\else\edef\valdecstr{\valdec}\fi%
  \ifdim\deltares pt>0pt%
    \pgfmathsetmacro{\deltaint}{int(min(100, abs(\deltares) * 3))}%
    \edef\deltacolor{deltagreen!\deltaint!white}%
    \begingroup\rmfamily\bfseries\small\expandafter\cellcolor\expandafter{\deltacolor}\valwhole.\valdecstr\endgroup%
  \else%
    \pgfmathsetmacro{\deltaint}{int(min(100, abs(\deltares) * 2))}%
    \edef\deltacolor{deltared!\deltaint!white}%
    \begingroup\rmfamily\bfseries\small\expandafter\cellcolor\expandafter{\deltacolor}\valwhole.\valdecstr\endgroup%
  \fi%
}

\newcommand{\testvaldeltabgpctul}[2]{%
  \pgfmathsetmacro{\deltares}{#2-#1}%
  \pgfmathtruncatemacro{\valwhole}{int(#2)}%
  \pgfmathtruncatemacro{\valdec}{int(round((#2-\valwhole)*100))}%
  \ifnum\valdec<10\edef\valdecstr{0\valdec}\else\edef\valdecstr{\valdec}\fi%
  \ifdim\deltares pt>0pt%
    \pgfmathsetmacro{\deltaint}{int(min(100, abs(\deltares) * 3))}%
    \edef\deltacolor{deltagreen!\deltaint!white}%
    \expandafter\cellcolor\expandafter{\deltacolor}\small\rmfamily\myul{\valwhole.\valdecstr}%
  \else%
    \pgfmathsetmacro{\deltaint}{int(min(100, abs(\deltares) * 2))}%
    \edef\deltacolor{deltared!\deltaint!white}%
    \expandafter\cellcolor\expandafter{\deltacolor}\small\rmfamily\myul{\valwhole.\valdecstr}%
  \fi%
}

\newcommand{\testvaldeltabgpctbful}[2]{%
  \pgfmathsetmacro{\deltares}{#2-#1}%
  \pgfmathtruncatemacro{\valwhole}{int(#2)}%
  \pgfmathtruncatemacro{\valdec}{int(round((#2-\valwhole)*100))}%
  \ifnum\valdec<10\edef\valdecstr{0\valdec}\else\edef\valdecstr{\valdec}\fi%
  \ifdim\deltares pt>0pt%
    \pgfmathsetmacro{\deltaint}{int(min(100, abs(\deltares) * 3))}%
    \edef\deltacolor{deltagreen!\deltaint!white}%
    \begingroup\rmfamily\bfseries\small\expandafter\cellcolor\expandafter{\deltacolor}\myul{\valwhole.\valdecstr}\endgroup%
  \else%
    \pgfmathsetmacro{\deltaint}{int(min(100, abs(\deltares) * 2))}%
    \edef\deltacolor{deltared!\deltaint!white}%
    \begingroup\rmfamily\bfseries\small\expandafter\cellcolor\expandafter{\deltacolor}\myul{\valwhole.\valdecstr}\endgroup%
  \fi%
}

\newcommand{\trainvaldeltabgpct}[2]{%
  \pgfmathsetmacro{\deltares}{#2-#1}%
  \pgfmathtruncatemacro{\valwhole}{int(#2)}%
  \pgfmathtruncatemacro{\valdec}{int(round((#2-\valwhole)*100))}%
  \ifnum\valdec<10\edef\valdecstr{0\valdec}\else\edef\valdecstr{\valdec}\fi%
  \ifdim\deltares pt>0pt%
    \pgfmathsetmacro{\deltaint}{int(min(100, abs(\deltares) * 20))}%
    \edef\deltacolor{deltagreen!\deltaint!white}%
    \expandafter\cellcolor\expandafter{\deltacolor}\small\rmfamily\valwhole.\valdecstr%
  \else%
    \pgfmathsetmacro{\deltaint}{int(min(100, abs(\deltares) * 6))}%
    \edef\deltacolor{deltared!\deltaint!white}%
    \expandafter\cellcolor\expandafter{\deltacolor}\small\rmfamily\valwhole.\valdecstr%
  \fi%
}

\newcommand{\trainvaldeltabgpctbf}[2]{%
  \pgfmathsetmacro{\deltares}{#2-#1}%
  \pgfmathtruncatemacro{\valwhole}{int(#2)}%
  \pgfmathtruncatemacro{\valdec}{int(round((#2-\valwhole)*100))}%
  \ifnum\valdec<10\edef\valdecstr{0\valdec}\else\edef\valdecstr{\valdec}\fi%
  \ifdim\deltares pt>0pt%
    \pgfmathsetmacro{\deltaint}{int(min(100, abs(\deltares) * 20))}%
    \edef\deltacolor{deltagreen!\deltaint!white}%
    \begingroup\rmfamily\bfseries\small\expandafter\cellcolor\expandafter{\deltacolor}\valwhole.\valdecstr\endgroup%
  \else%
    \pgfmathsetmacro{\deltaint}{int(min(100, abs(\deltares) * 6))}%
    \edef\deltacolor{deltared!\deltaint!white}%
    \begingroup\rmfamily\bfseries\small\expandafter\cellcolor\expandafter{\deltacolor}\valwhole.\valdecstr\endgroup%
  \fi%
}

\newcommand{\trainvaldeltabgpctul}[2]{%
  \pgfmathsetmacro{\deltares}{#2-#1}%
  \pgfmathtruncatemacro{\valwhole}{int(#2)}%
  \pgfmathtruncatemacro{\valdec}{int(round((#2-\valwhole)*100))}%
  \ifnum\valdec<10\edef\valdecstr{0\valdec}\else\edef\valdecstr{\valdec}\fi%
  \ifdim\deltares pt>0pt%
    \pgfmathsetmacro{\deltaint}{int(min(100, abs(\deltares) * 20))}%
    \edef\deltacolor{deltagreen!\deltaint!white}%
    \expandafter\cellcolor\expandafter{\deltacolor}\small\rmfamily\myul{\valwhole.\valdecstr}%
  \else%
    \pgfmathsetmacro{\deltaint}{int(min(100, abs(\deltares) * 6))}%
    \edef\deltacolor{deltared!\deltaint!white}%
    \expandafter\cellcolor\expandafter{\deltacolor}\small\rmfamily\myul{\valwhole.\valdecstr}%
  \fi%
}

\newcommand{\trainvaldeltabgpctbful}[2]{%
  \pgfmathsetmacro{\deltares}{#2-#1}%
  \pgfmathtruncatemacro{\valwhole}{int(#2)}%
  \pgfmathtruncatemacro{\valdec}{int(round((#2-\valwhole)*100))}%
  \ifnum\valdec<10\edef\valdecstr{0\valdec}\else\edef\valdecstr{\valdec}\fi%
  \ifdim\deltares pt>0pt%
    \pgfmathsetmacro{\deltaint}{int(min(100, abs(\deltares) * 20))}%
    \edef\deltacolor{deltagreen!\deltaint!white}%
    \begingroup\rmfamily\bfseries\small\expandafter\cellcolor\expandafter{\deltacolor}\myul{\valwhole.\valdecstr}\endgroup%
  \else%
    \pgfmathsetmacro{\deltaint}{int(min(100, abs(\deltares) * 6))}%
    \edef\deltacolor{deltared!\deltaint!white}%
    \begingroup\rmfamily\bfseries\small\expandafter\cellcolor\expandafter{\deltacolor}\myul{\valwhole.\valdecstr}\endgroup%
  \fi%
}

\newcommand{\chartdeltabgpct}[2]{%
  \pgfmathsetmacro{\deltares}{#2-#1}%
  \pgfmathtruncatemacro{\valwhole}{int(#2)}%
  \pgfmathtruncatemacro{\valdec}{int(round((#2-\valwhole)*100))}%
  \ifnum\valdec<10\edef\valdecstr{0\valdec}\else\edef\valdecstr{\valdec}\fi%
  \ifdim\deltares pt>0pt%
    \pgfmathsetmacro{\deltaint}{int(min(100, abs(\deltares) * 1.8))}%
    \edef\deltacolor{deltagreen!\deltaint!white}%
    \expandafter\cellcolor\expandafter{\deltacolor}\small\rmfamily\valwhole.\valdecstr%
  \else%
    \pgfmathsetmacro{\deltaint}{int(min(100, abs(\deltares) * 2))}%
    \edef\deltacolor{deltared!\deltaint!white}%
    \expandafter\cellcolor\expandafter{\deltacolor}\small\rmfamily\valwhole.\valdecstr%
  \fi%
}

\newcommand{\chartdeltabgpctbf}[2]{%
  \pgfmathsetmacro{\deltares}{#2-#1}%
  \pgfmathtruncatemacro{\valwhole}{int(#2)}%
  \pgfmathtruncatemacro{\valdec}{int(round((#2-\valwhole)*100))}%
  \ifnum\valdec<10\edef\valdecstr{0\valdec}\else\edef\valdecstr{\valdec}\fi%
  \ifdim\deltares pt>0pt%
    \pgfmathsetmacro{\deltaint}{int(min(100, abs(\deltares) * 1.8))}%
    \edef\deltacolor{deltagreen!\deltaint!white}%
    \begingroup\rmfamily\bfseries\small\expandafter\cellcolor\expandafter{\deltacolor}\valwhole.\valdecstr\endgroup%
  \else%
    \pgfmathsetmacro{\deltaint}{int(min(100, abs(\deltares) * 2))}%
    \edef\deltacolor{deltared!\deltaint!white}%
    \begingroup\rmfamily\bfseries\small\expandafter\cellcolor\expandafter{\deltacolor}\valwhole.\valdecstr\endgroup%
  \fi%
}

\newcommand{\chartdeltabgpctul}[2]{%
  \pgfmathsetmacro{\deltares}{#2-#1}%
  \pgfmathtruncatemacro{\valwhole}{int(#2)}%
  \pgfmathtruncatemacro{\valdec}{int(round((#2-\valwhole)*100))}%
  \ifnum\valdec<10\edef\valdecstr{0\valdec}\else\edef\valdecstr{\valdec}\fi%
  \ifdim\deltares pt>0pt%
    \pgfmathsetmacro{\deltaint}{int(min(100, abs(\deltares) * 1.8))}%
    \edef\deltacolor{deltagreen!\deltaint!white}%
    \expandafter\cellcolor\expandafter{\deltacolor}\small\rmfamily\myul{\valwhole.\valdecstr}%
  \else%
    \pgfmathsetmacro{\deltaint}{int(min(100, abs(\deltares) * 2))}%
    \edef\deltacolor{deltared!\deltaint!white}%
    \expandafter\cellcolor\expandafter{\deltacolor}\small\rmfamily\myul{\valwhole.\valdecstr}%
  \fi%
}

\newcommand{\chartdeltabgpctbful}[2]{%
  \pgfmathsetmacro{\deltares}{#2-#1}%
  \pgfmathtruncatemacro{\valwhole}{int(#2)}%
  \pgfmathtruncatemacro{\valdec}{int(round((#2-\valwhole)*100))}%
  \ifnum\valdec<10\edef\valdecstr{0\valdec}\else\edef\valdecstr{\valdec}\fi%
  \ifdim\deltares pt>0pt%
    \pgfmathsetmacro{\deltaint}{int(min(100, abs(\deltares) * 1.8))}%
    \edef\deltacolor{deltagreen!\deltaint!white}%
    \begingroup\rmfamily\bfseries\small\expandafter\cellcolor\expandafter{\deltacolor}\myul{\valwhole.\valdecstr}\endgroup%
  \else%
    \pgfmathsetmacro{\deltaint}{int(min(100, abs(\deltares) * 2))}%
    \edef\deltacolor{deltared!\deltaint!white}%
    \begingroup\rmfamily\bfseries\small\expandafter\cellcolor\expandafter{\deltacolor}\myul{\valwhole.\valdecstr}\endgroup%
  \fi%
}

\newcommand{\charttrainvaldeltabgpct}[2]{%
  \pgfmathsetmacro{\deltares}{#2-#1}%
  \pgfmathtruncatemacro{\valwhole}{int(#2)}%
  \pgfmathtruncatemacro{\valdec}{int(round((#2-\valwhole)*100))}%
  \ifnum\valdec<10\edef\valdecstr{0\valdec}\else\edef\valdecstr{\valdec}\fi%
  \ifdim\deltares pt>0pt%
    \pgfmathsetmacro{\deltaint}{int(min(100, abs(\deltares) * 5.2))}%
    \edef\deltacolor{deltagreen!\deltaint!white}%
    \expandafter\cellcolor\expandafter{\deltacolor}\small\rmfamily\valwhole.\valdecstr%
  \else%
    \pgfmathsetmacro{\deltaint}{int(min(100, abs(\deltares) * 2))}%
    \edef\deltacolor{deltared!\deltaint!white}%
    \expandafter\cellcolor\expandafter{\deltacolor}\small\rmfamily\valwhole.\valdecstr%
  \fi%
}

\newcommand{\charttrainvaldeltabgpctbf}[2]{%
  \pgfmathsetmacro{\deltares}{#2-#1}%
  \pgfmathtruncatemacro{\valwhole}{int(#2)}%
  \pgfmathtruncatemacro{\valdec}{int(round((#2-\valwhole)*100))}%
  \ifnum\valdec<10\edef\valdecstr{0\valdec}\else\edef\valdecstr{\valdec}\fi%
  \ifdim\deltares pt>0pt%
    \pgfmathsetmacro{\deltaint}{int(min(100, abs(\deltares) * 5.2))}%
    \edef\deltacolor{deltagreen!\deltaint!white}%
    \begingroup\rmfamily\bfseries\small\expandafter\cellcolor\expandafter{\deltacolor}\valwhole.\valdecstr\endgroup%
  \else%
    \pgfmathsetmacro{\deltaint}{int(min(100, abs(\deltares) * 2))}%
    \edef\deltacolor{deltared!\deltaint!white}%
    \begingroup\rmfamily\bfseries\small\expandafter\cellcolor\expandafter{\deltacolor}\valwhole.\valdecstr\endgroup%
  \fi%
}

\newcommand{\charttrainvaldeltabgpctul}[2]{%
  \pgfmathsetmacro{\deltares}{#2-#1}%
  \pgfmathtruncatemacro{\valwhole}{int(#2)}%
  \pgfmathtruncatemacro{\valdec}{int(round((#2-\valwhole)*100))}%
  \ifnum\valdec<10\edef\valdecstr{0\valdec}\else\edef\valdecstr{\valdec}\fi%
  \ifdim\deltares pt>0pt%
    \pgfmathsetmacro{\deltaint}{int(min(100, abs(\deltares) * 5.2))}%
    \edef\deltacolor{deltagreen!\deltaint!white}%
    \expandafter\cellcolor\expandafter{\deltacolor}\small\rmfamily\myul{\valwhole.\valdecstr}%
  \else%
    \pgfmathsetmacro{\deltaint}{int(min(100, abs(\deltares) * 2))}%
    \edef\deltacolor{deltared!\deltaint!white}%
    \expandafter\cellcolor\expandafter{\deltacolor}\small\rmfamily\myul{\valwhole.\valdecstr}%
  \fi%
}

\newcommand{\charttrainvaldeltabgpctbful}[2]{%
  \pgfmathsetmacro{\deltares}{#2-#1}%
  \pgfmathtruncatemacro{\valwhole}{int(#2)}%
  \pgfmathtruncatemacro{\valdec}{int(round((#2-\valwhole)*100))}%
  \ifnum\valdec<10\edef\valdecstr{0\valdec}\else\edef\valdecstr{\valdec}\fi%
  \ifdim\deltares pt>0pt%
    \pgfmathsetmacro{\deltaint}{int(min(100, abs(\deltares) * 5.2))}%
    \edef\deltacolor{deltagreen!\deltaint!white}%
    \begingroup\rmfamily\bfseries\small\expandafter\cellcolor\expandafter{\deltacolor}\myul{\valwhole.\valdecstr}\endgroup%
  \else%
    \pgfmathsetmacro{\deltaint}{int(min(100, abs(\deltares) * 2))}%
    \edef\deltacolor{deltared!\deltaint!white}%
    \begingroup\rmfamily\bfseries\small\expandafter\cellcolor\expandafter{\deltacolor}\myul{\valwhole.\valdecstr}\endgroup%
  \fi%
}

%% file: sections/1_intro.tex
\section{Introduction}
\label{sec:intro}

\input{figures/1_teaser}

Large language models have grown from single-turn responders into agents that plan, call external tools, and interact with their environment \citep{schick2023toolformer,wang2023voyager,shen2023hugginggpt}. More recently, they begin to organize one another \citep{hong2024metagpt,zhuge2024gptswarm}. Given a task and a pool of available specialist agents, an agent now writes the division of work itself, deciding which agent takes each part and in what order the parts run \citep{zhang2025aflow,zhou2026mass,yue2026autoras}.
What such a flow contributes is structure: \textit{the work is divided explicitly, each part goes to an agent suited to it, and progress accumulates step by step rather than resting on one long attempt}.

Yet the decisions such an agent makes are narrower than they appear. How finely the task is divided is fixed in advance, such as by a template, a prompt, or a chosen number of steps. Where it does adapt, it adapts to the task alone \citep{liu2025selectthendecompose,madhwal-2026-decomposed-prompting,su2026daao}. What it would cost the available agents to carry out one division rather than another seldom enters.
Where nothing available fits, the work is either forced onto another agent or a new one is created, and neither choice is priced against the other \citep{zhang2025swarmagentic,wu2025kamac,shang2025agentsquare}.
Once the flow runs, the evidence execution produces goes unused: identifying which step goes wrong is treated as needing a reference answer, a graded outcome, or a trained assessor. The first two do not exist while the flow is running, and the third requires supervision the task alone cannot provide. Methods that dispense with all three optimize only once the run has finished, by replaying the completed trace to locate the failure or rerunning the whole flow for a better answer \citep{chae2025webshepherd,lee2026prints,lin2026reflectinterventionsupportederrorattribution}.
Meanwhile, when something goes wrong, the unit of fix is the whole flow, \textit{e.g.}, re-executed, re-searched, or its constructor re-trained. Thus, improvement is priced at the size of the workflow rather than of the fault, and every step that already succeeds is paid for twice \citep{hwang2026agenticmontecarlosimulating,li2026agentflow,dang2025puppeteer}.
Together, these leave both challenges of \S\ref{sec:prelim} open: \textit{building a workflow well under uncertainty and improving one at lower cost}.
Worse still, neither is put at risk: \textit{the tasks these systems are measured on are largely ones a single competent agent already handles} \citep{zhou2026mass,sun2026collaborationtaxllmmultiagent,choi2025debateorvote}.

\ours answers these in two stages (\S\ref{sec:method}), pricing every choice in one currency that needs no labels: \textit{how well an agent's competence meets what a part of the work demands, weighed against how much that agent takes to run}.
In the first stage, \coalesce decomposes the task to its finest parts and find the proper granularity through coalescing (\S\ref{subsec:method:decomposition}).
In the second stage, \ours performs \textit{in-flow dynamic optimization} by turning the same currency on the flow as it runs, identifying and correcting subtasks locally without losing the execution progress (\S\ref{subsec:method:inflowop}). One label-free cost thus serves the workflow both as it is built and as it runs.
To sum up, our main contributions are:
\begin{oursenumerate}
  \item \textbf{\coalesce: the decomposition granularity is priced, not presumed.} A cost-driven bidirectional workflow construction approach (Alg.~\ref{alg:coalesce}) over the agents on hand settles how finely the task is divided, enabling agent creation only when no available agent clears a competence bar (\S\ref{subsec:method:decomposition}).

  \item \textbf{\ours: optimization in flow, without supervision.} We propose \ours (Alg.~\ref{alg:overview}), a two-stage training-free framework that leverages \coalesce to build workflows and optimizes them in-flow (Alg.~\ref{alg:inflowop}), without any reference answer, graded outcome, or trained assessor (\S\ref{subsec:method:inflowop}).

  \item \textbf{\braid: a \underline{b}enchmark for \underline{r}easoning over \underline{a}gent-\underline{i}nterdependent \underline{d}ivisions.} Workflow-level tasks are those a single agent cannot solve effectively (\S\ref{appendix:sec:braid}). This is the premise that motivates workflows in the first place, yet one that existing benchmarks seldom satisfy (\S\ref{appendix:subsec:braid:necessity}). We introduce \braid (\S\ref{sec:braid}), a workflow-level benchmark that requires multi-agent coordinations beyond single-agent competence.

  \item \textbf{Evaluation: gains across domains and backbones.} Our evaluation across eight domains, six backbones, and four workflow constructors shows that \ours improves over a single agent from $+7.15\%$ to $+11.97\%$ under a matched turn budget, and outperforms workflows baselines by up to $+9.64\%$ (\S\ref{sec:exps}).
\end{oursenumerate}

%% file: figures/1_teaser.tex
\begin{wrapfigure}[21]{r}{0.6\textwidth}
    \small
    \centering
    \vspace{-42pt}
    
    \includegraphics[width=0.6\textwidth]{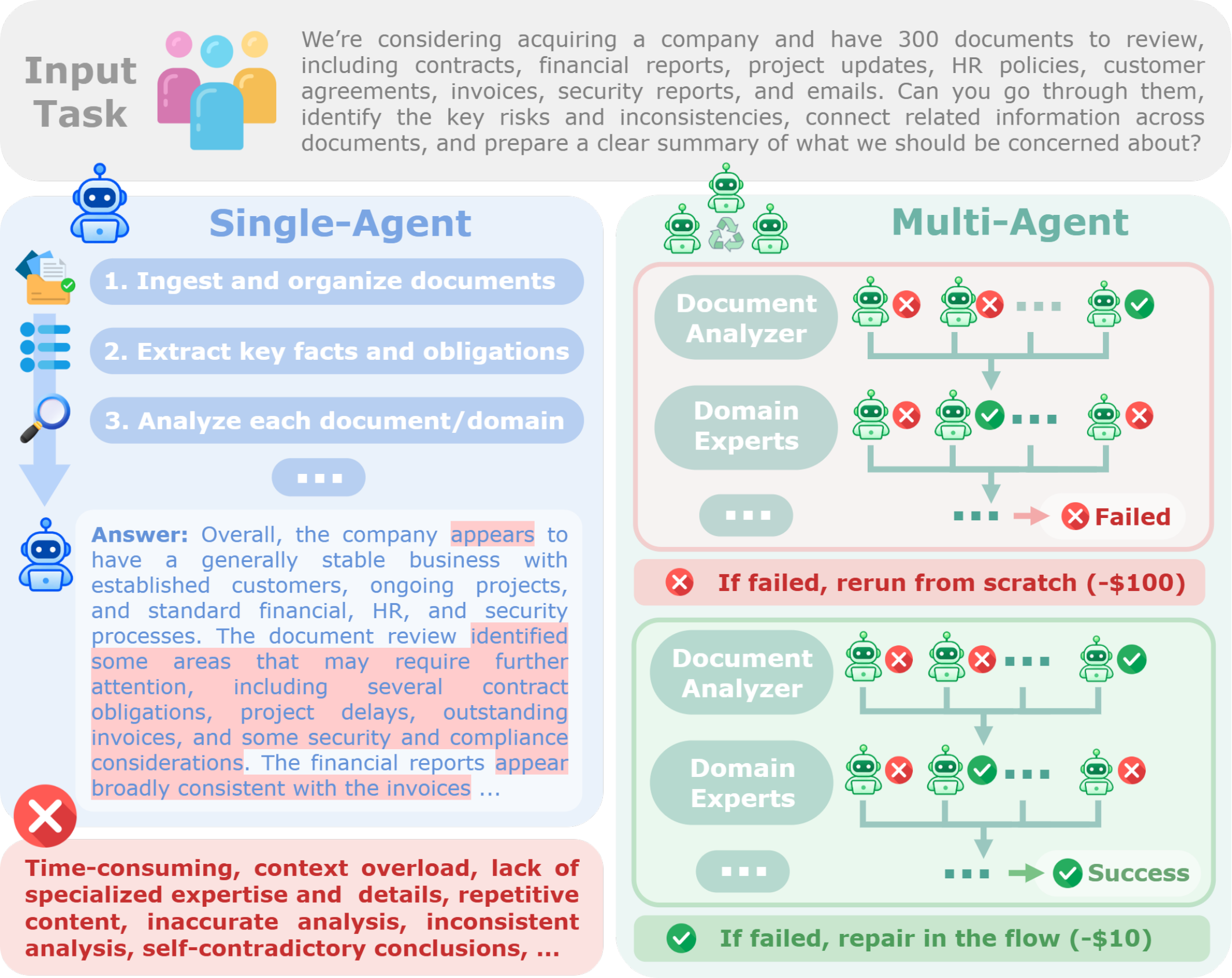}
    
    \vspace{-6pt}
    
    \caption{\textbf{\ours for In-Flow Optimization.} Complex tasks challenge single agents with long-context reasoning, while multi-agent systems introduce execution failures and coordination overhead. \ours optimize agent behaviors in-flow, enabling efficient collaboration at lower cost.}
    \label{fig:teaser}

    \vspace{-6pt}
\end{wrapfigure}

%% file: sections/2_prelim.tex
\section{Preliminaries}
\label{sec:prelim}

\label{subsec:prelim:paradigm}

When tackling complex tasks that cannot be resolved by a single agent, we build multi-agent workflows:

\textbf{The problem.}
A task $Q$ is addressed using a pool of specialist agents $\mathcal{A}=\{a_1,\dots,a_m\}$. An LLM-powered workflow constructor $\mathcal{D}$ produces a workflow $W=\mathcal{D}(Q,\mathcal{A})$ that orchestrates agents from $\mathcal{A}$. The workflow $W$ is then executed to evaluate its performance $\mathrm{Perf}(W)\in[0,1]$ against the reference answer of $Q$. The objective is to maximize $\mathrm{Perf}$ at minimal cost via the flow $\mathcal{D} \rightarrow W \rightarrow \mathrm{Perf}$. However, this is affected by (1) $\mathcal{D} \rightarrow W$: how $W$ is \textit{constructed}, and (2) $W \rightarrow \mathrm{Perf}$: how $\mathrm{Perf}$ is \textit{optimized} (\S\ref{sec:method}).

\begin{hlbox}{Problem Formulation}
\textbf{Pool-Aware Workflow Construction:} Given a task and a pool of specialist agents, automatically assemble a multi-agent workflow that solves the task both \textit{effectively} and \textit{efficiently}.

\begin{oursenumerate}
    \item \textbf{\textit{Effective to solve} (\textit{reliability}).} A workflow is expected to reliably produce the correct final answer. When composing many agents, its reliability is bounded by each agent's progress.
    \item \textbf{\textit{Efficient to run} (\textit{execution}).} A more complex task calls for a more complex workflow, accordingly more subtasks, agents, and latency to run. So the complexity of a workflow should be matched to the task, spending no more than the task demands.
    \item \textbf{\textit{Efficient to improve} (\textit{optimization}).} When the workflow falls short, improving it should be cheap, without the time and compute of full re-execution, re-search, or model training.
\end{oursenumerate}
\end{hlbox}

\label{subsec:prelim:build}

\textbf{Building a Workflow under Uncertainty.}
%
Specialist agents in the pool differ markedly in competence, thus binding a subtask to an ill-suited agent is a primary source of failure \citep{liu2025selectthendecompose,madhwal-2026-decomposed-prompting,su2026daao}.
A constructor $\mathcal{D}$ that composes agents without regard to their fit therefore produces unreliable workflows.
%
%
On the other hand, the agent pool is also finite and may hold no agent well-suited to its assigned work. This induces new agent \textit{creation} rather than forcing a poor fit, while at the added cost of specifying and instantiating it \citep{zhang2025swarmagentic,wu2025kamac,shang2025agentsquare}.
%
%
Nevertheless, even a well-matched agent is not guaranteed to succeed: whether it completes its assigned work correctly is \textit{uncertain}, and that uncertainty surfaces only once the workflow runs, which is why fault attribution is studied on completed executions \citep{chae2025webshepherd,lee2026prints,lin2026reflectinterventionsupportederrorattribution}. Execution alone is not conclusive either, as a step that satisfies what it declares can still mislead the steps that follow (\S\ref{appendix:subsec:inflowop:outcome_only}).
%
%
Collectively, these raise the first core challenge of workflow construction: \textit{how to build a good workflow under uncertainty?}

\label{subsec:prelim:optimize}

\textbf{The Cost of Workflow Optimization.}
Once a workflow runs, execution finally reveals which steps falter. However, acting on such evidence is expensive.
%
%
The naive remedy is to re-execute the workflow or re-search over a set of candidate workflows, while discarding everything already computed and progressed \citep{hwang2026agenticmontecarlosimulating,li2025encompass,yoon2026patplanningaftertrialefficienttesttime}. An increment of quality thereby charges additional full execution (Fig.~\ref{fig:exp_cost}). Worse still, it presumes a supervision signal absent at inference: no ground-truth answer is available to say \textit{what} went wrong or \textit{where} to improve.
%
%
On the other hand, a more targeted way is post-training, guiding $\mathcal{D}$ to learn on success cases via supervised finetuning or improve via reward signals through reinforcement learning \citep{li2026agentflow,peng2026hiper,nielsen2026conductor,dang2025puppeteer}. Whereas the time and computation cost of data preparation and model training can go far beyond test-time re-execution.
%
%
These studies reveal the second core challenge on workflow optimization: \textit{how to improve a workflow at lower cost?}


\begin{hlbox}{Agent Card}
Every agent is represented to the estimator by a card, the sole evidence an atom is scored against (1) \emph{declared competence}: a written brief of what the agent is designed for, including its role, domain, and the skills and tools it commands; and (2) \emph{self-evolving empirical profile}: the record an agent accumulates through execution, profiling which kinds of work it has realized successfully and where it has faltered.
So a declared competence is progressively complemented by one that is observed.
\end{hlbox}

%% file: sections/3_method.tex
\section{InFlowOp: the Flow as Built, the Flow as Run}
\label{sec:method}


\input{figures/2_inflowop}

\textit{How to build a reliable workflow, and optimize it at low cost} (\S\ref{sec:prelim})?
Rather than relying on costly training or sampling, we cast both questions as minimizing a single measurable, annotation-free \emph{surrogate}: the workflow's \emph{cost}, which couples its reliability of success with its latency to run.
As such, we minimize this cost in two stages (\S\ref{appendix:subsec:inflowop:stages}): (1) \emph{decomposition-aware workflow construction} (\S\ref{subsec:method:decomposition}) constructs a low-cost workflow before execution, and (2) \emph{in-flow dynamic optimization} (\S\ref{subsec:method:inflowop}) \textit{tempers} it adaptively during execution.

\subsection{Bidirectional Decomposition-Aware Workflow Construction}
\label{subsec:method:decomposition}

\textbf{Decomposition.}
A complex task seldom fits any single agent. Instead, its parts demand different competences that no one agent commands (\S\ref{subsec:prelim:build}). Therefore, we \emph{decompose} the task into subtasks, each narrow enough to hand to a well-fit agent and, being narrower, more reliably completed. A workflow is thus a triple $W=\langle S,A,G\rangle$, where $S=\{s_1,\dots,s_{n'}\}$ is a set of subtasks, $A$ denotes an assignment binding a subtask $s_i$ to an agent $a(s_i)$, and $G$ represents a dependency graph, which is a directed acyclic graph (DAG) with edges linking subtasks' nodes. Thus, the \emph{critical path} of $G$, \textit{i.e.}, its longest chain of dependent subtasks, sets the workflow's latency.
Consequently, the right granularity is a question of \emph{cost}: finer decomposition narrows each subtask and raises per-subtask reliability, but multiplies agents to run with increased latency cost and the points at which the workflow may fail with increased reliability cost. This motivates our \coalesce: \textit{bidirectional cost-driven atomic coalescing} (Alg.~\ref{alg:coalesce}).

\input{algorithms/overview}

\textbf{Cost matrix.}
Scoring Eq.~\ref{eq:workflow_cost} is based on the reliability cost $C(\tau,a)$ of realizing an atom $\tau$ with an agent $a$. We compute it \emph{without labels} by \textit{matching} each agent's declared competence and self-evolving empirical profile against what that atom demands, via a rubric semantic estimator that scores how well a supplied text satisfies a demanded one: $\hat p(\tau,a)=\mathrm{match}(\textit{demand},\textit{supply})=\mathrm{match}(\tau,a)\in[0,1]$. Running over all atoms and agents, it constructs the \emph{cost matrix}, with rows as atoms and columns as agents.
%
An agent \emph{covers} an atom when its cost clears a competence bar $\bar C$, so the matrix also exposes where the pool falls short:
\begin{equation}
\rho(\tau)=\one\big[\min_{a\in\mathcal{A}}C(\tau,a)>\bar C\big]
\label{eq:coverage}
\end{equation}
the \emph{coverage residual} of $\tau$: $\rho(\tau)=1$ marks an atom no pooled agent can be trusted with, which triggers new agent creation at penalty $\gamma$. With \textit{dynamic matrix} enabled, each created agent joins the matrix as a new column that every atom can be matched to.
As such, we obtain the \emph{cost matrix} with latencies $\ell(\tau,a)$ to guide \coalesce in finding proper granularity for constructing better workflows (\S\ref{subsec:method:decomposition}).

\begin{hlbox}{Taxonomy of Cost}
With various kinds of costs incurred across each workflow's construction, execution, and optimization stages (\S\ref{sec:prelim}), we organize them into one systematic cost taxonomy (\S\ref{appendix:subsec:inflowop:cost_taxonomy}):

\begin{oursenumerate}
    \item \textbf{\textit{Reliability cost}:} $R$ (Eq.~\ref{eq:reliability}). How much a workflow risks failing, \textit{i.e.}, the price of entrusting each subtask to an agent that may not realize it correctly.
    \item \textbf{\textit{Latency cost}:} $L$ (Eq.~\ref{eq:latency}). How much a workflow spends to run, \textit{i.e.}, the time and compute its execution consumes.
    \item \textbf{\textit{Creation cost}:} $\lvert\mathcal{A}_+\rvert$ (Eq.~\ref{eq:workflow_cost}). The price of creating a new agent when the pool holds none that fits, under the uncertainty of its true competence.
    \item \textbf{\textit{Workflow cost}:} $\mathrm{Cost}(W)$ (Eq.~\ref{eq:workflow_cost}). The overall cost of a \textit{constructed} workflow, jointly balancing reliability, latency, and creation, exactly what our construction method aims to minimize (Eq.~\ref{eq:construction_objective}).
    \item \textbf{\textit{Optimization cost}.} The price of improving a workflow, \textit{e.g.}, the effort of re-execution, re-search, re-sampling, model training, \textit{etc.}, separate from the workflow’s own cost.
\end{oursenumerate}

\end{hlbox}

\textbf{Cost-driven coalescing.}
Not every atom is needed: we keep only $\nu\subseteq\mathcal{T}$, those from which the goal is reachable over $G_{\mathcal{T}}$, and discard the rest (Alg.~\ref{alg:coalesce}). \coalesce then proceeds over \emph{partial coalescings} $P$, each placing a prefix of $\nu$ in topological order. Extending $P$ by its next atom $\tau_i$ admits two moves: $\textsc{Open}(\tau_i)$ starts a new subtask, and $\textsc{Merge}(\tau_i,s)$ folds $\tau_i$ into an open subtask $s$, so a workflow is compacted only where merging pays. A partial coalescing is dropped once it can no longer improve on the best complete one found, so the optimum always survives (\S\ref{appendix:subsec:inflowop:stages}). \coalesce returns the least-cost coalescing $P^{\star}$ within $B$ expansions, and $\textsc{Lift}$ then carries the atom dependencies $G_{\mathcal{T}}$ to the subtasks of $P^{\star}$, inducing the subtask graph $G$.

\textbf{Objective.}
Decomposition-aware construction thus reduces to one optimization over decompositions $S$, assignments $A$, and the dependency graph $G$ they induce, for the least-cost workflow:
\begin{equation}
W^\star=\arg\min_{\langle S,A,G\rangle}\ \mathrm{Cost}(W)=\arg\min_{\langle S,A,G\rangle}\ R(W)+\beta\,L(W)+\gamma\,\lvert\mathcal{A}_+(W)\rvert
\label{eq:construction_objective}
\end{equation}
where $G$ is the dependency graph induced by the decomposition $S$, and $\mathcal{A}_+(W)$ denotes the agents created for $W$. \coalesce (Alg.~\ref{alg:coalesce}) solves it through bidirectional cost-driven atomic coalescing, scoring every candidate pairing via cost matrix and atom coalescing to descend $\mathrm{Cost}(W)$.
This cost prices structures \coalesce emits via meta primitives (\S\ref{appendix:subsec:inflowop:meta_primitives}).

\begin{hlbox}{Bidirectional Cost-Driven Atomic Coalescing}
Rather than fixed granularity, we let \textit{cost} steer the appropriate granularity in two directions:

\begin{oursenumerate}
    \item \textbf{\emph{Top-down atomic decomposition}:} the task $Q$ is decomposed into its finest atoms $\mathcal{T}=\{\tau_1,\dots,\tau_n\}$, each individually reliable.
    \item \textbf{\emph{Bottom-up cost-driven coalescing}:} guided by \textit{cost matrix} (\S\ref{subsec:method:decomposition}), these atoms are \emph{coalesced} into $n'\le n$ subtasks $S$, each bound to its lowest-cost agent, either existing in pool or newly created, merging only where a coarser subtask lowers the overall cost.
\end{oursenumerate}

Granularity is therefore an \emph{outcome} of cost-driven coalescing rather than a fixed choice, adapting to each task's requirements and complexity (\coalesce, Alg.~\ref{alg:coalesce}).
\end{hlbox}

\subsection{In-Flow Dynamic Optimization}
\label{subsec:method:inflowop}

A constructed workflow is only a \emph{prediction} of what is expected to work (\S\ref{subsec:prelim:build}). Execution is therefore the first place a workflow's real weaknesses become visible, while also potentially expensive to act on if acting means re-running or re-searching the workflow in full (\S\ref{subsec:prelim:optimize}). Instead, we optimize \emph{in flow}: a fault is \textit{caught} as it arises, execution \textit{pauses} at the subtask it arises in, the workflow is \textit{tempered} there locally, and execution \textit{resumes} over everything already achieved.

\begin{hlbox}{In-Flow Dynamic Optimization}
Execution proceeds subtask by subtask. Whenever a realized subtask is found faulty, we \emph{temper} the workflow there, and execution resumes, rather than restarting it:

\begin{oursenumerate}
    \item \textbf{\emph{Detect}:} match each realized subtask against the contract its decomposition declared, using only what execution makes observable \textit{without} reference.
    \item \textbf{\emph{Pause}:} halt at the subtask that falters, not at the workflow it belongs to, so every progress already achieved stays intact.
    \item \textbf{\emph{Temper}:} correct the fault along a ladder of moves ordered by \textit{cost matrix} (\S\ref{subsec:method:decomposition}), committing a change only once it demonstrably holds.
    \item \textbf{\emph{Resume}:} re-run only the tempered subtask, retaining every result already computed.
\end{oursenumerate}
\end{hlbox}

\textbf{Detection.}
%
%
%
%
%
\coalesce does not merely coalesce atoms into subtasks. Instead, it establishes each subtask with the \textit{contract} its atoms declare, containing the input conditions $\sigma^i_{\mathrm{in}}$ that $s_i$ requires, and the output conditions $\sigma^i_{\mathrm{out}}$ it is expected to deliver. This contract is what makes a fault observable \textit{without} supervision by stating what \emph{should} hold without knowing the answer. Detection then reuses \textit{cost matrix} (\S\ref{subsec:method:decomposition}) and applies retrospectively: where construction matches an atom's demands against an agent's card, in-flow detection matches each declared condition against the work actually realized, the observed input $\mathcal{I}_i$ and output $\mathcal{O}_i$ of $s_i$:
\begin{equation}
u(\sigma\,,x)=1-\mathrm{match}(\sigma\,,x)\in[0,1],
\qquad
\sigma \text{ is \emph{unmet} iff } u(\sigma\,,x)\ge\bar u
\label{eq:credit}
\end{equation}
with $x=\mathcal{I}_i$ for input conditions and $x=\mathcal{O}_i$ for output ones. The resulting \emph{credit matrix} is the retrospective counterpart of \textit{cost matrix}.
%
Which side breaches \emph{blames} the fault, and the blame is what a correction respects:
\begin{equation}
\mathrm{blame}(s_i)=
\begin{cases}
\textit{decomposition}, & \text{if some } \sigma\in\sigma^i_{\mathrm{in}} \text{ is unmet}\\[2pt]
\textit{assignment}, & \text{else if some } \sigma\in\sigma^i_{\mathrm{out}} \text{ is unmet}\\[2pt]
\textit{none}, & \text{otherwise}
\end{cases}
\label{eq:fault_type}
\end{equation}
An unmet \emph{output} condition means the subtask is well posed but its agent underdelivered. An unmet \emph{input} condition means the subtask is not given what it needs, which no choice of agent can undo but \coalesce. Alongside this \textit{semantic} signal, we also retain a \emph{causal} signal: a subtask that yields no usable output is faulty by observation and needs no matching.

\textbf{Cost-ordered ladder.}
We define two moves to correct a workflow at a faulty subtask, and the cost model of \S\ref{subsec:method:decomposition} prices both. \textsc{Re-assign} leaves $S$ and $G$ untouched and updates the agent, perturbing a single term of Eq.~\ref{eq:reliability}. \textsc{Re-decompose} updates $s_i$ by a finer sub-DAG, changing $S$ and $G$ together by adding subtasks, dependencies, and possibly a created agent, and paying a fresh run of \coalesce besides. Pricing the two in that shared currency orders them:
\begin{equation}
\textsc{Re-assign}\ \prec\ \textsc{Re-decompose}
\label{eq:ladder}
\end{equation}
which defines the \emph{cost-ordered ladder}: spend the cheapest move the fault admits, and climb only once that move is exhausted. The blame (Eq.~\ref{eq:fault_type}) fixes the entry rung: a fault blamed on \emph{assignment} enters at \textsc{Re-assign}, one blamed on \emph{decomposition} at \textsc{Re-decompose}.

\textbf{Temper.}
At the \textsc{Re-assign} rung, the \textit{cost matrix} answers directly: the next-best agent is the same column-wise minimum construction takes, now restricted to agents not yet observed to fail on $s_i$,
\begin{equation}
a'(s_i)=\arg\min_{a\,\in\,(\mathcal{A}\cup\mathcal{A}_+)\setminus\mathcal{X}(s_i)} C\big(s_i,a\big)
\label{eq:reassign}
\end{equation}
where $\mathcal{X}(s_i)$ collects those attempted and ruled out. \textsc{Re-assign} is committed only if $s_i$ executes \emph{and} its unmet conditions clear under Eq.~\ref{eq:credit}; otherwise it is rolled back so that no unverified change survives. When the candidates are exhausted, the evidence has ruled out the assignment as the cause and the ladder climbs: \coalesce is re-invoked on $s_i$ locally, \emph{under its declared contract} $(\sigma^i_{\mathrm{in}},\sigma^i_{\mathrm{out}})$, and the resulting sub-DAG is spliced in its place.
Every move is budgeted, $B_{\mathrm{ra}}$ for \textsc{Re-assign} and $B_{\mathrm{rd}}$ for \textsc{Re-decompose}, and a subtask that exhausts its budget reverts to its original realization: \textit{tempering can improve a workflow, but never leaves it worse than the one construction produces.}

\textbf{Scoped resumption.}
Execution pauses at $s_i$, and tempering it invalidates only the subtasks that depend on it, so every result computed outside that region remains valid. Execution therefore resumes over the \emph{affected closure} alone:
\begin{equation}
I(s_i)=\{s_i\}\cup\mathrm{Desc}(s_i)
\label{eq:closure}
\end{equation}
where $\mathrm{Desc}(s_i)$ are the subtasks reachable from $s_i$ in $G$, and replay cached results for every subtask outside $I(s_i)$. The price of an optimization step is thus the cost of the \emph{local} correction only, bounded by $I(s_i)$ rather than by $|S|$. An optional global localization is elaborated in \S\ref{appendix:subsec:inflowop:global}.

\textbf{Objective.}
In-flow optimization descends the same cost from the constructed workflow $W^{\star}$ (Eq.~\ref{eq:construction_objective}) under evidence:
\begin{equation}
W^{\dagger}=\arg\min_{W'\,\in\,\mathcal{N}^{\star}(W^{\star})}\ \widetilde{\mathrm{Cost}}(W'),
\qquad
\tilde C(s,a)=
\begin{cases}
\infty, & (s,a)\in\mathcal{X}\\[2pt]
C(s,a), & \text{otherwise}
\end{cases}
\label{eq:inflowop_objective}
\end{equation}
where $\mathcal{N}^{\star}(W^{\star})$ is the set of workflows reachable from $W^{\star}$ by ladder moves (Eq.~\ref{eq:ladder}), each confined to a closure $I(s_i)$ (Eq.~\ref{eq:closure}); $\mathcal{X}$ collects the pairings observed to breach their contracts (Eq.~\ref{eq:reassign}); and $\widetilde{\mathrm{Cost}}$ is Eq.~\ref{eq:workflow_cost} evaluated with $\tilde C$ in place of $C$.

%% file: figures/2_inflowop.tex
\begin{figure*}[!t]

    \small
    \centering
    \includegraphics[width=1.0\textwidth]{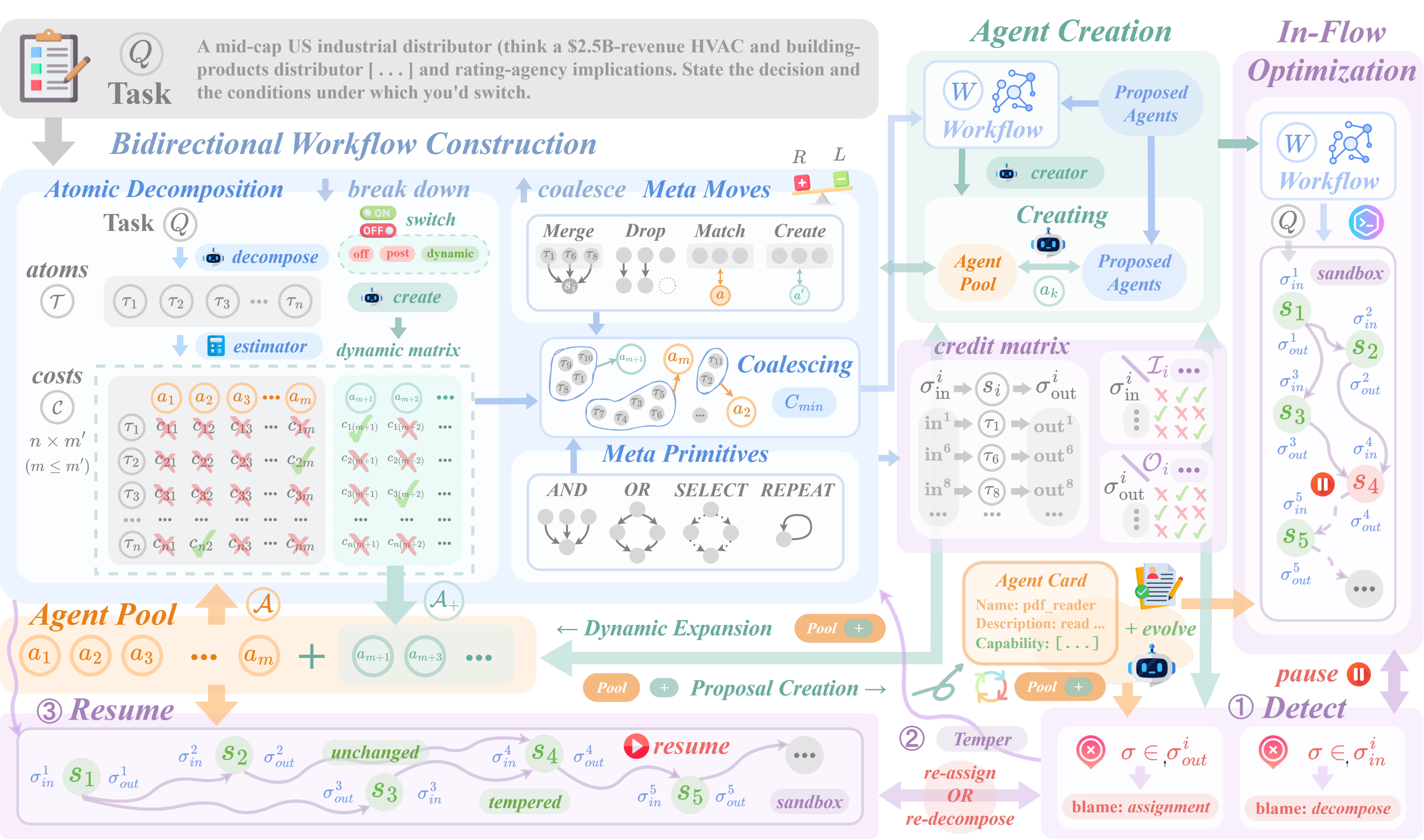}
    
    \vspace{-6pt}
    
    \caption{\textbf{Bidirectional Workflow Construction and In-Flow Optimization.} Our \textit{bidirectional decomposition-aware workflow construction} (\S\ref{subsec:method:decomposition}) builds a workflow $W$ at the proper granularity with minimal cost $C_\textit{min}$ in two directions: \textit{top-down decomposition} and \textit{bottom-up coalescing} (Alg.~\ref{alg:coalesce}). \textit{In-flow dynamic optimization} (\S\ref{subsec:method:inflowop}, Alg.~\ref{alg:inflowop}) then optimizes $W$ by retrospectively applying the cost matrix to construct the credit matrix, fixing faulty intermediate steps in flow without reference answer. The constructed workflow is executed in the \textit{sandbox}, where \ours optimizes it as the execution proceeds (Alg.~\ref{alg:overview}).}
    \label{fig:inflowop}

    \vspace{-3pt}

\end{figure*}

%% file: algorithms/overview.tex
\begin{algorithm}[t]

\caption{\ours$(Q,\mathcal{A})$: The Flow as Built, the Flow as Run (\S\ref{sec:method}, \S\ref{appendix:subsec:inflowop:stages})}
\label{alg:overview}

\begin{algorithmic}[1]
\Require task $Q$; agent pool $\mathcal{A}$
\Statex \textcolor{commentcolor}{\textit{\textbf{One estimator serves both stages:} $\mathrm{match}(\textit{demand},\textit{supply})$ prices pairings into $C$, and scores declared conditions against realized states in flow. Stage II inherits $C$ the workflow with contracts from Stage I.}}
\Ensure the final answer $y$ and the workflow $W^{\dagger}$ that produced it

\Statex \textcolor{commentcolor}{\textit{\textbf{Stage I: Workflow Construction} (\S\ref{subsec:method:decomposition}; Alg.~\ref{alg:coalesce})}}
\State $C(s,a)\gets-\log\mathrm{match}(s,a)$;\quad $\rho(s)\gets\one\big[\min_a C(s,a)>\bar C\big]$ \Comment{\textit{cost matrix}; where the pool falls short (Eq.~\ref{eq:reliability},~\ref{eq:coverage})}
\State $W^{\star},\{(\sigma^{i}_{\mathrm{in}},\sigma^{i}_{\mathrm{out}})\}\gets\textsc{Coalesce}\big(Q,\mathcal{A}\mid C\big)$ \Comment{$C$ settles granularity and agent assignment (Eq.~\ref{eq:construction_objective})}

\Statex \textcolor{commentcolor}{\textit{\textbf{Stage II: In-Flow Optimization} (\S\ref{subsec:method:inflowop}; Alg.~\ref{alg:inflowop})}}
\ForAll{$s_i\in W^{\star}$} \Comment{in topological order}
   \State $(\mathcal{I}_i,\mathcal{O}_i)\gets\textsc{Execute}\big(s_i,a(s_i)\big)$
   \State $u(\sigma,x)\gets1-\mathrm{match}(\sigma,x)$ \Comment{\textit{credit matrix}: $x=\mathcal{I}_i$ for $\sigma^{i}_{\mathrm{in}}$ and $x=\mathcal{O}_i$ for $\sigma^{i}_{\mathrm{out}}$ (Eq.~\ref{eq:credit})}
   \State $b\gets\mathrm{blame}(s_i)$ \Comment{which side of the contract is breached (Eq.~\ref{eq:fault_type})}
   \If{$b\neq\textit{none}$}
      \State $\textsc{Pause}(s_i)$ \Comment{halt at the fault, not at the flow}
      \State $W'\gets\textsc{Temper}\big(W,s_i,b\mid C,\sigma^{i}\big)$ \Comment{$b=\textit{assignment}$ (Eq.~\ref{eq:reassign}) or $b=\textit{decomposition}$ (Eq.~\ref{eq:ladder})}
      \State $W\gets W'$ \textbf{ iff } $u(\sigma,\cdot)<\bar u\ \ \forall\sigma$ \Comment{else roll back: never worse than built}
      \State $\textsc{Resume}\big(I(s_i)\big)$ \Comment{only the fault is paid for (Eq.~\ref{eq:closure})}
   \EndIf
\EndFor

\State \Return $y,\ W^{\dagger}$ \Comment{Eq.~\ref{eq:inflowop_objective}}
\end{algorithmic}

\end{algorithm}

%% file: sections/4_braid.tex
\section{\braid: Benchmark Reasoning over Agent-Interdependent Divisions}
\label{sec:braid}

\input{tables/exp_main}

\textbf{Benchmark construction.}
A workflow earns its necessity only where a single agent cannot handle the task alone, yet workflows are largely measured on suites one competent agent already solves (\S\ref{appendix:subsec:braid:necessity}). \braid closes this gap by composing each task at different \textit{complexity levels} $k$ via two composition methods: \textit{distributed} composition gives every question its own gold source, while \textit{anchored} composition shares one gold source across all provides sources. We enforce four quality-control criteria, and a task failing any of them is never written (\S\ref{appendix:subsec:braid:construction}). Built on $10$ public datasets, \braid spans $19$ evaluation-only arms across $8$ domains (Tab.~\ref{tab:appendix:braid_statistics}).

\textbf{Evaluation metrics.}
\braid grades every question separately according to its answer type, so the accuracy of a task is the mean score over its $k$ questions and the accuracy of a \braid dataset is the mean over its tasks (\S\ref{appendix:subsec:braid:eval_metrics}). Our rubric evaluation metrics cover answer types spanning \textit{numeric}, \textit{mathematical}, \textit{textual}, \textit{multiple-choice}, \textit{list}, and \textit{coding} answers.

%% file: tables/exp_main.tex
\begin{table*}[t]

\centering
\caption{\textbf{Main Results.} Mean accuracy (\%) per domain on \braid, averaged over the arms each domain holds and over both complexity levels (\S\ref{sec:braid}). \textit{All} (\%) averages over the $16$ arms of these five domains.}
\label{tab:exp_main}

\vspace{3pt}
\small
\renewcommand{\arraystretch}{1.2}
\newcommand{\blockgap}{\noalign{\vskip 2pt}}
\setlength{\tabcolsep}{2.5pt}

\begin{tabularx}{\linewidth}{@{}>{\bfseries}l*{7}{>{\centering\arraybackslash}X}@{}}
\toprule
\multicolumn{1}{@{}c}{\textbf{Method}} & \cellcolor{blue_bg}\textcolor{blue_text}{\textbf{Doc}} & \cellcolor{blue_bg}\textcolor{blue_text}{\textbf{Fin}} & \cellcolor{blue_bg}\textcolor{blue_text}{\textbf{Chart}} & \cellcolor{blue_bg}\textcolor{blue_text}{\textbf{Math}} & \cellcolor{blue_bg}\textcolor{blue_text}{\textbf{Phys}} & \cellcolor{blue_bg}\textcolor{blue_text}{\textbf{All}} & \cellcolor{blue_bg}\textcolor{blue_text}{\textbf{$\Delta_{\text{acc}}$}} \\
\midrule

\rowcolor{gray_bg}
\multicolumn{8}{@{}c@{}}{\textcolor{gray_text}{\textbf{Baseline: Single LLM (\textit{no tools or skills})}}} \\
\blockgap
Qwen3.5-4B & 5.64 & 0.52 & 2.15 & 0.00 & 0.00 & 1.74 & -- \\
Qwen3.5-9B & 5.95 & 1.64 & 4.58 & 3.31 & 0.54 & 3.23 & -- \\
GPT-5-mini & 4.02 & 3.34 & 15.75 & 10.05 & 11.79 & 8.85 & -- \\
GPT-5.4-mini & 9.67 & 1.97 & 17.12 & 12.32 & 10.82 & 10.59 & -- \\
GPT-5.4 & 10.43 & 2.54 & 14.07 & 42.73 & 13.02 & 18.62 & -- \\
GPT-5.6-luna & 12.98 & 3.87 & 25.20 & 66.66 & 18.84 & 28.25 & -- \\
\blockgap

\rowcolor{gray_bg}
\multicolumn{8}{@{}c@{}}{\textcolor{gray_text}{\textbf{Baseline: Single Agent (\textit{with full pool of tools and skills})}}} \\
\blockgap
Qwen3.5-4B & 12.51 & 4.83 & 22.93 & 23.48 & 4.78 & 13.66 & \gaindeltaflat{1.74}{13.66} \\
Qwen3.5-9B & 11.26 & 20.27 & 25.88 & 30.38 & 4.81 & 17.38 & \gaindeltaflat{3.23}{17.38} \\
GPT-5-mini & 23.02 & 3.54 & 22.83 & 19.13 & 10.00 & 16.33 & \gaindeltaflat{8.85}{16.33} \\
GPT-5.4-mini & 25.44 & 1.38 & 24.91 & 21.02 & 10.38 & 17.49 & \gaindeltaflat{10.59}{17.49} \\
GPT-5.4 & 33.68 & 25.67 & 42.17 & 35.56 & 10.52 & 27.21 & \gaindeltaflat{18.62}{27.21} \\
GPT-5.6-luna & 41.11 & 49.03 & 58.36 & 58.20 & 14.94 & 41.99 & \gaindeltaflat{28.25}{41.99} \\
\blockgap

\rowcolor{blue_bg}
\multicolumn{8}{@{}c@{}}{\textcolor{blue_text}{\textbf{\textsc{InFlowOp} (\textit{ours})}}} \\
\blockgap
Qwen3.5-4B & \gaincell{5.64}{14.91} & \gaincell{0.52}{12.87} & \gaincell{2.15}{32.45} & \gaincell{0.00}{38.48} & \gaincell{0.00}{7.17} & \gaincell{1.74}{20.81} & \gaindeltaflat{1.74}{20.81} \\
Qwen3.5-9B & \gaincell{5.95}{14.81} & \gaincell{1.64}{25.20} & \gaincell{4.58}{41.28} & \gaincell{3.31}{42.65} & \gaincell{0.54}{9.80} & \gaincell{3.23}{25.13} & \gaindeltaflat{3.23}{25.13} \\
GPT-5-mini & \gaincell{4.02}{26.71} & \gaincell{3.34}{19.23} & \gaincell{15.75}{37.67} & \gaincell{10.05}{40.42} & \gaincell{11.79}{12.44} & \gaincell{8.85}{27.01} & \gaindeltaflat{8.85}{27.01} \\
GPT-5.4-mini & \gaincell{9.67}{28.96} & \gaincell{1.97}{21.14} & \gaincell{17.12}{40.55} & \gaincell{12.32}{43.09} & \gaincell{10.82}{14.93} & \gaincell{10.59}{29.46} & \gaindeltaflat{10.59}{29.46} \\
GPT-5.4 & \gaincell{10.43}{45.24} & \gaincell{2.54}{41.41} & \gaincell{14.07}{55.61} & \gaincell{42.73}{50.52} & \gaincell{13.02}{14.85} & \gaincell{18.62}{38.52} & \gaindeltaflat{18.62}{38.52} \\
GPT-5.6-luna & \gaincell{12.98}{42.22} & \gaincell{3.87}{56.95} & \gaincell{25.20}{70.23} & \gaincell{66.66}{74.33} & \gaincell{18.84}{17.34} & \gaincell{28.25}{49.37} & \gaindeltaflat{28.25}{49.37} \\

\bottomrule
\end{tabularx}

\end{table*}

%% file: sections/5_exps.tex
\section{Experiments}
\label{sec:exps}


\input{figures/exp_domains}

\subsection{Experiment Setup}
\label{subsec:exps:setup}

\textbf{Setup.}
We compare six systems under the same pool of agents. (1) \textit{Single LLM} answers a task in one model call, without any tools or skills\footnote{The single-LLM baseline scores near zero on several domains because every input is provided as a file: without tools or skills, it cannot access them, so only answers from what it already knows.}. (2) \textit{Single agent} answers with the full tools and skills provided. The remaining four construct workflows: (3) \textit{greedy-search} and (4) \coalesce build workflows respectively, and our comparison between them isolates what the decomposition alone contributes. (5) \coalesce $+$ \textit{in-flow} adds the tempering of \S\ref{subsec:method:inflowop}, and (6) the full method further scores each agent on the self-evolving profile, rather than on its card alone.

\textbf{Models.}
We evaluate two open-source models, \texttt{Qwen3.5} (\texttt{4B} and \texttt{9B})~\citep{qwen2026qwen35omnitechnicalreport}, together with four close-source models, \texttt{GPT-5-mini}~\citep{openai2025gpt5mini}, \texttt{GPT-5.4-mini}~\citep{openai2026gpt54mini}, \texttt{GPT-5.4}~\citep{openai2026gpt54}, \texttt{GPT-5.6-luna}~\citep{openai2026gpt56luna}.

\textbf{Data.}
We evaluate on \braid (\S\ref{sec:braid}), covering 19 evaluation arms over 14 \braid datasets adapted from 10 public datasets across 8 domains (Tab.~\ref{tab:appendix:braid_statistics}). Each is constructed at two complexity levels and two compositions a dataset supports\footnote{Due to high API cost, we evaluate \texttt{GPT-5.4-mini}, \texttt{GPT-5.4}, and \texttt{GPT-5.6-luna} on the $5$-domain subset of \braid (Tab.~\ref{tab:exp_main}). Fig.~\ref{fig:exp_domains} gives all eight domains for the three models evaluated on the whole benchmark.}.


\subsection{What the Flow Earns and What Earns the Flow}
\label{subsec:exps:main}

\textbf{Consistent gains across domains and backbones.}
\ours improves every backbone over both single-LLM and single-agent baselines (Tab.~\ref{tab:exp_main}), achieving gains up to $+21.90\%$ over the single LLM baseline, and up to $+11.97\%$ over the single agent baseline with matched compute budget (\S\ref{appendix:sec:implementation}). What the two baselines earn depends on the backbone, as the single agent gains from $+6.90\%$ to $+14.15\%$ over the single LLM. On the other hand, what \ours earns varies by $\leq3.74\%$ across six backbones.
The gain also exceeds what a larger backbone delivers: under \ours, \texttt{Qwen3.5-4B} reaches $20.81\%$, above the single-agent accuracy of \texttt{Qwen3.5-9B}, \texttt{GPT-5-mini}, and \texttt{GPT-5.4-mini}. Meanwhile, the strongest backbone \texttt{GPT-5.6-luna} also improves from $41.99\%$ to $49.37\%$. Across domains, the gain is largest on charts, mathematics, and finance, while smallest on physics, where all methods stay $\leq17.34\%$ under any backbone. Also, the same improvement trends hold consistently on slides, science, and code, the three additional domains for the three backbones evaluated on the whole benchmark (Fig.~\ref{fig:exp_domains}).

\textbf{Gains come not from dividing the task, but from how it is divided.}
Comparing against the single agent baseline evaluated on the same \braid arms (Tab.~\ref{tab:exp_modules}), a workflow constructed greedily obtains $15.49\%$, falling below the single-agent's $17.38\%$. Therefore, dividing a task is not by itself what earns the gain. Deciding the division by cost reverses that: \coalesce reaches $22.21\%$, $+6.72\%$ above greedy search and $+4.83\%$ above the single-agent baseline. In-flow tempering adds a further $+1.59\%$, and scoring each agent on the evolving profile improves $+1.33\%$ more. Each module therefore earns its place, and the two (3)-(4) optimizing in-flow earn less than the one deciding the workflow before it runs.

\input{tables/exp_modules}

\subsection{What Improves the Flow and What Weighs on It}
\label{subsec:exps:ablation}

\input{figures/exp_abl_estimator_scope}

\textbf{Cost is better computed than judged.}
\ours prices the cost matrix with a rubric estimator to match atom demands against agent capabilities (\S\ref{subsec:method:decomposition}). This cost matrix can also be priced by an LLM that judges matching scores among agents for each atom in one pass. We ablate our rubric estimator against the LLM counterpart. As shown in Fig.~\ref{fig:exp_abl_estimator_scope} (\textit{left}), the LLM estimator loses $-3.27\%$ points on \texttt{Qwen3.5-4B} and $-3.85\%$ on \texttt{Qwen3.5-9B}, leaving \texttt{Qwen3.5-9B} below the $22.21\%$ it reaches under \coalesce alone (Tab.~\ref{tab:exp_modules}). Both estimators are given the same agent cards, evolving profiles, and atom demands. Therefore, what separates them is not what they know, but whether their estimated cost can effectively guide atom matching and coalescing to construct better workflows.

\textbf{Dynamic cost matrix that updates during \coalesce guides workflow construction better.}
Different from dynamic cost matrix that adaptively update with newly created agents, a static cost matrix is constructed once and then stays fixed. This leaves every new agent \coalesce creates outside the matching with other atoms in static matrix.
As shown in Fig.~\ref{fig:exp_abl_construction}, \ours with dynamic matrix outperforms the static matrix counterpart by $+3.15\%$ on \texttt{Qwen3.5-4B} and $+2.55\%$ on \texttt{Qwen3.5-9B}. Therefore, a cost matrix dynamic evolving with the agent pool is able to more effectively guide \coalesce to find more proper granularities.

\input{figures/exp_abl_construction}

\textbf{Pools that dynamically grow build better workflows.}
The two pools differ where no available agent fits an atom well: a static agent pool settles for the closest agent it happens to hold, while a dynamic agent pool creates the specialist that atom demands. As shown in Fig.~\ref{fig:exp_abl_construction}, \ours with a dynamic pool outperforms its static counterpart by $+4.49\%$ on \texttt{Qwen3.5-4B} and $+3.29\%$ on \texttt{Qwen3.5-9B}. Therefore, the pool sets a ceiling that no coalescing can lift: where no available agent covers an atom, dividing the task differently only relocates the shortfall, whereas creating the missing specialist removes it.

\input{figures/exp_abl_profile}

\textbf{Evolving profiles help only when they remember success.}
An agent can be scored on its agent card alone, or on the card together with the evolving profile it accumulates. As shown in Fig.~\ref{fig:exp_abl_profile}, a profile evolving within a single run gains $+2.54\%$, $+1.33\%$, and $+2.23\%$ over the agent card alone on \texttt{Qwen3.5-4B}, \texttt{Qwen3.5-9B} and \texttt{GPT-5.4-mini}, respectively. When continuously evolving it into later runs, \ours gains a further $+1.21\%$, $+1.19\%$, and $+1.30\%$ when only successful ones are kept. Applying all to profiles instead reaches $17.34\%$, $21.52\%$, and $26.53\%$ respectively for three backbones, below the agent card alone on all three backbones. This indicates that what makes a profile worth evolving is evidence of what an agent can do: a success tells the estimator what to trust that agent with, whereas a failure marks only where it once fell short, and prices the agent no better than no profile provided.

\textbf{Global localization rarely fires, and loses accuracy when it does.}
Beyond local in-flow optimization, can a workflow be improved further by a global pass over the same cost matrix (\S\ref{appendix:subsec:inflowop:global})? To investigate this, we compare accuracy under \textit{local}-only in-flow optimization against \textit{local}$+$\textit{global}. As shown in Fig.~\ref{fig:exp_abl_estimator_scope} (\textit{right}), enabling the global pass beside the local one yields $-1.63\%$ on \texttt{Qwen3.5-4B} and $-1.41\%$ on \texttt{Qwen3.5-9B}. This reveals that a breached contract already points at what to temper, and that tempering beyond it costs more than it recovers. This finding is also the empirical evidence for our configuration, under which in-flow optimization stays local in every other experiment.

%% file: figures/exp_domains.tex
\begin{figure*}[t]
    \vspace{0pt}
    
    \small
    \centering

    \includegraphics[width=\textwidth]{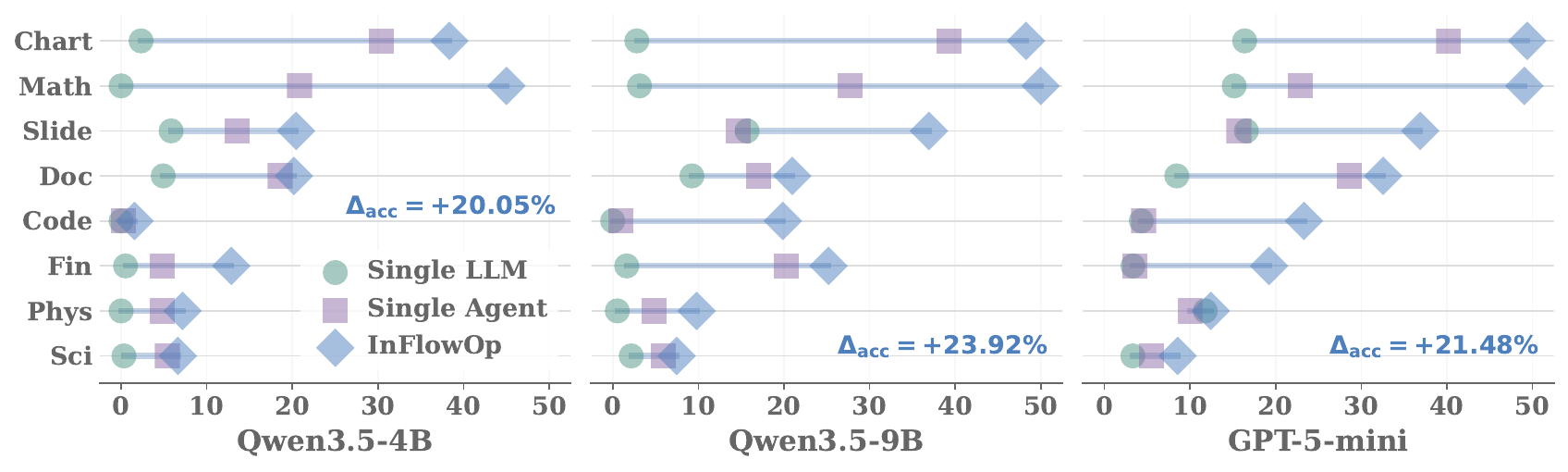}

    \vspace{-12pt}

    \caption{\textbf{Evaluation on All Eight Domains of \braid.} Mean accuracy (\%) per domain for the three models evaluated on the whole benchmark. Each row shows a domain under the two baselines and \ours (\S\ref{subsec:exps:setup}). The horizontal axis is accuracy (\%). Tab.~\ref{tab:exp_main} covers the five domains every model is evaluated on.}
    \label{fig:exp_domains}

    \vspace{10pt}

\end{figure*}

%% file: tables/exp_modules.tex
\begin{table*}[t]
\vspace{-6pt}

\centering
\caption{\textbf{What Each Module Contributes.} Mean accuracy (\%) of \texttt{Qwen3.5-9B} on \braid. \ours is assembled one module at a time in (2)-(4): (1) a workflow built greedily, (2) a workflow built with \coalesce, (3) the same \coalesce-built workflow with in-flow tempering, and (4) the full method (\S\ref{sec:method}, \S\ref{appendix:subsec:inflowop:stages}, Alg.~\ref{alg:overview}).}
\label{tab:exp_modules}

\vspace{3pt}
\small
\renewcommand{\arraystretch}{1.2}
\newcommand{\blockgap}{\noalign{\vskip 2pt}}
\setlength{\tabcolsep}{2.5pt}

\begin{tabularx}{\linewidth}{@{}>{\bfseries}l*{7}{>{\centering\arraybackslash}X}@{}}
\toprule
\multicolumn{1}{@{}c}{\textbf{Method}} & \cellcolor{blue_bg}\textcolor{blue_text}{\textbf{Doc}} & \cellcolor{blue_bg}\textcolor{blue_text}{\textbf{Fin}} & \cellcolor{blue_bg}\textcolor{blue_text}{\textbf{Chart}} & \cellcolor{blue_bg}\textcolor{blue_text}{\textbf{Math}} & \cellcolor{blue_bg}\textcolor{blue_text}{\textbf{Phys}} & \cellcolor{blue_bg}\textcolor{blue_text}{\textbf{All}} & \cellcolor{blue_bg}\textcolor{blue_text}{\textbf{$\Delta_{\text{acc}}$}} \\
\midrule
Single Agent & 11.26 & 20.27 & 25.88 & 30.38 & 4.81 & 17.38 & -- \\
\blockgap
greedy-search & \gaincell{11.26}{8.85} & \gaincell{20.27}{13.46} & \gaincell{25.88}{17.63} & \gaincell{30.38}{33.00} & \gaincell{4.81}{4.55} & \gaincell{17.38}{15.49} & \gaindeltaflat{17.38}{15.49} \\
\coalesce & \gaincell{11.26}{12.19} & \gaincell{20.27}{21.25} & \gaincell{25.88}{38.55} & \gaincell{30.38}{39.15} & \gaincell{4.81}{7.61} & \gaincell{17.38}{22.21} & \gaindeltaflat{17.38}{22.21} \\
\coalesce $+$ in-flow & \gaincell{11.26}{13.47} & \gaincell{20.27}{22.75} & \gaincell{25.88}{39.99} & \gaincell{30.38}{40.86} & \gaincell{4.81}{9.48} & \gaincell{17.38}{23.80} & \gaindeltaflat{17.38}{23.80} \\
\ours & \gaincell{11.26}{14.81} & \gaincell{20.27}{25.20} & \gaincell{25.88}{41.28} & \gaincell{30.38}{42.65} & \gaincell{4.81}{9.80} & \gaincell{17.38}{25.13} & \gaindeltaflat{17.38}{25.13} \\
\bottomrule
\end{tabularx}

\vspace{6pt}

\end{table*}

%% file: figures/exp_abl_estimator_scope.tex
\begin{wrapfigure}[15]{r}{0.5\textwidth}
    \vspace{-13pt}
    
    \small
    \centering

    \includegraphics[width=0.98\linewidth]{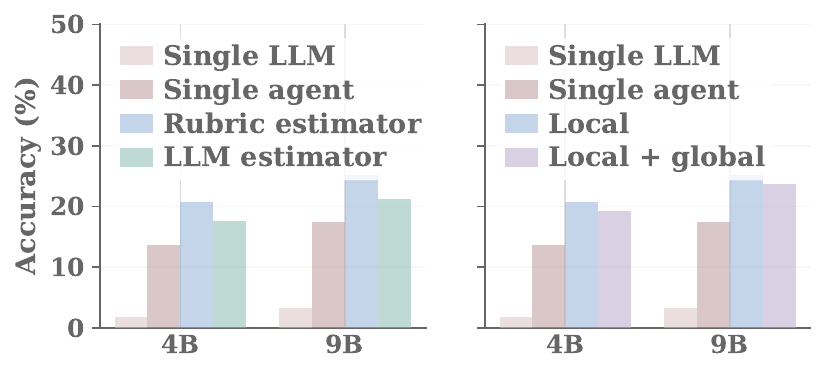}

    \vspace{-12pt}

    \caption{\textbf{How Fit Is Scored, and Where a Fault Is Sought.} Mean accuracy (\%) on \texttt{Qwen3.5-4B} and \texttt{Qwen3.5-9B}. \textit{Left}: using rubric estimator against an LLM for cost estimation. \textit{Right}: in-flow optimization looking where a contract breaches, against the same with the global pass enabled beside it (\S\ref{appendix:subsec:inflowop:global}).}
    \label{fig:exp_abl_estimator_scope}

\end{wrapfigure}

%% file: figures/exp_abl_construction.tex
\begin{wrapfigure}[12]{r}{0.5\textwidth}
    \vspace{0pt}
    
    \small
    \centering

    \includegraphics[width=0.98\linewidth]{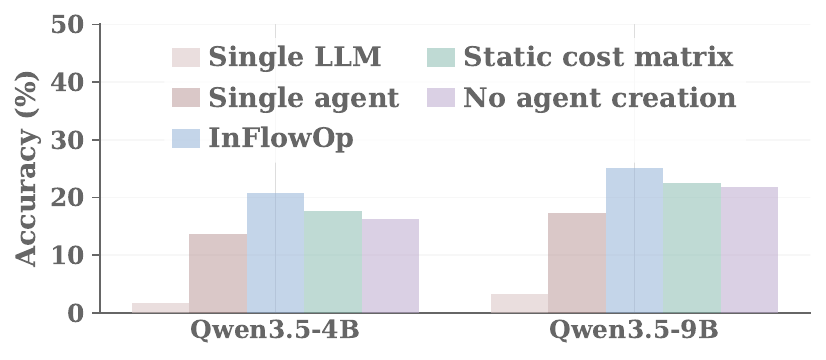}

    \vspace{-12pt}

    \caption{\textbf{Construction Earns Its Two Freedoms.} Mean accuracy (\%) over \texttt{Qwen3.5} (\texttt{4B} and \texttt{9B}) among static cost matrix, dynamic cost matrix, and static pool.}
    \label{fig:exp_abl_construction}

\end{wrapfigure}

%% file: figures/exp_abl_profile.tex
\begin{wrapfigure}[18]{r}{0.5\textwidth}
    \vspace{6pt}
    
    \small
    \centering

    \includegraphics[width=0.98\linewidth]{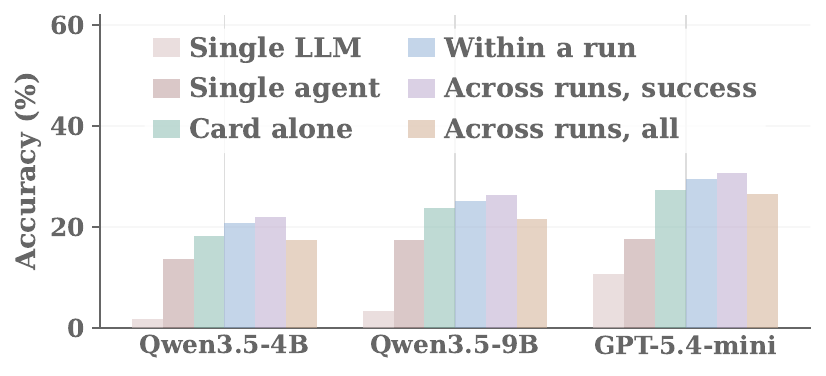}

    \vspace{-12pt}

    \caption{\textbf{A Profile Is Worth What It Remembers.} Mean accuracy (\%) over an agent scored (1) on its card alone, (2) on a profile evolving within a run, and (3) on a profile evolving across runs, keeping either successful ones only or all of them.}
    \label{fig:exp_abl_profile}

\end{wrapfigure}

%% file: sections/8_end.tex
\vspace{-6pt}

\section{Conclusions}
\label{sec:conclusion}

\vspace{-6pt}


In this work, we present \ours (\S\ref{sec:method}), a two-stage framework (\S\ref{appendix:subsec:inflowop:stages}) of \textit{bidirectional decomposition-aware construction} (\S\ref{subsec:method:decomposition}) and \textit{in-flow dynamic optimization} (\S\ref{subsec:method:inflowop}). Given the challenge of workflow-level evaluation, we contribute \braid, a benchmark whose tasks demand multi-agent coordination beyond single-agent capability (\S\ref{sec:braid}). We evaluate \ours across eight domains and six backbones (\S\ref{subsec:exps:setup}) against single-model and workflow baselines (\S\ref{subsec:exps:main}) and ablate its design choices (\S\ref{subsec:exps:ablation}).

%% file: sections/appendix.tex
\clearpage
\appendix

\input{sections/6_results}

\input{sections/7_related_work}

\section{Limitations}
\label{appendix:sec:limitations}

\ours is evaluated on \braid, whose tasks span eight domains and are graded against a reference answer for each question (\S\ref{sec:braid}). Tasks settled by no checkable answer, such as open-ended generation or long-horizon interaction with a stateful environment, lie outside this study.
In future work, we plan to extend \ours to such settings and different domains. We also plan to investigate workflows whose agents act on an environment that changes as they run.

\section{Where a Workflow Becomes Necessary: \braid for Workflow-Level Evaluation}
\label{appendix:sec:braid}

What a workflow contributes is coordination, especially how the work is divided among agents and how those divisions depend on one another. Therefore, measuring the contribution of a workflow requires complex tasks whose demands exceed what one agent can meet. However, existing benchmarks rarely supply such task complexity: their tasks are simple enough for one competent agent to solve alone. Consequently, a workflow evaluated on them is measured on tasks that require neither task decomposition nor multi-agent coordination beyond single-agent competence (\S\ref{sec:intro}, \ref{subsec:related_work:evaluation}, \ref{appendix:subsec:braid:necessity}). To close this gap, we propose \braid, a benchmark for workflow-level agent-interdependent reasoning (\S\ref{sec:braid}). Concretely, \braid composes every task through two composition methods under quality control (\S\ref{appendix:subsec:braid:construction}), and measures the effectiveness of a workflow by the mean accuracy over the questions each task comprises (\S\ref{appendix:subsec:braid:eval_metrics}).

\subsection{Why Workflow-Level Evaluation Is Necessary}
\label{appendix:subsec:braid:necessity}

\paragraph{When a workflow earns its necessity.}
A workflow is worth building only where a single agent cannot effectively carry the task alone, yet workflows are seldom evaluated on tasks that reach such complexity (\S\ref{sec:intro}, \ref{subsec:related_work:evaluation}). As shown in Fig.~\ref{fig:appendix:braid_necessity}, we conduct a preliminary test on this gap, comparing a single agent solving the task in one pass with a workflow an LLM constructs for the same task: when evaluated on the original data, the single agent is more accurate than the workflow ($-12.02\%$) with notably lower cost ($-6.91$). Therefore, coordination is not a necessity for simple tasks, and running multi-agent workflows on them can further incur several times the cost. On the other hand, with \braid composition (\S\ref{appendix:subsec:braid:construction}), the ordering reverses: the single agent degrades significantly on such complex tasks, while the workflow, running under a matched compute budget, showcases the strength of task decomposition and multi-agent coordination with better performance. Workflow-level task complexity is therefore a necessity for workflow evaluation, and the objective \braid targets.

\input{figures/appendix_braid_necessity}

\paragraph{Why workflow-level evaluation is necessary.}
The benchmarks a multi-agent workflow is usually measured on sit on the left side of each subplot (Fig.~\ref{fig:appendix:braid_necessity}: left \textit{original} half of both subplots): a single competent agent is already able to solve their tasks, which therefore demand neither task decomposition nor coordination among agents. As a result, evaluating a workflow on them cannot effectively measure how a workflow performs and what it actually contributes. Instead, reliable workflow-level evaluation requires tasks whose complexity demands multi-agent coordination beyond single-agent capability.

\subsection{Benchmarking Workflow-Level Evaluation}
\label{appendix:subsec:braid:construction}

\braid constructs workflow-level tasks by composing each task with $k$ questions and providing extra sources beyond those its questions require, without stating which source serves which question. We use $k$ to denote the \textit{complexity level} of a task. Answering one task therefore demands both halves of what a workflow is for: \textit{dividing} the task, and reaching its answers through coordinated \textit{routing}. Here, we define \textit{routing} as locating the source a question is answered from and reasoning over it to that question's answer. We use \textit{solver} to denote an evaluated system in general: a single LLM, an agent, or a multi-agent workflow that receives the task and produces its answers.

\paragraph{Composition methods.}
A task is composed in one of two methods: \textit{distributed} or \textit{anchored}. A \textit{source} here is one document or image provided with the task, however many files it takes to provide it, and the source a question is answered from is its \textit{gold source}. They differ in where the gold sources of a task's questions sit and therefore in what the \textit{division} and the \textit{routing} each demand:

\begin{oursenumerate}
  \item \textbf{\textit{Distributed}: every question has its own source.} Each \textit{distributed} task takes $k$ questions that are each answered from a different source. \braid provides those $k$ sources, and additionally inputs $(k-1)$ more sources that answer none of the questions, which we call \textit{distractors}. As a result, a \textit{distributed} task inputs $(2k-1)$ sources in total.

  \item \textbf{\textit{Anchored}: all questions share one source.} An \textit{anchored} task takes $k$ questions that are all answered from one source. \braid provides that source, and additionally inputs $(k-1)$ distractors alongside it. As a result, an \textit{anchored} task inputs $k$ sources in total.  
\end{oursenumerate}

\paragraph{Source selection.}
Distractors come from inside \braid, and the kind of input decides which extra sources supply them. A dataset whose inputs are documents draws its distractors from its own split, so a distractor is the same kind of document as the gold source and differs only in what it contains. A dataset whose inputs are images draws from a different dataset of the same kind, which keeps routing real via strict quality control while ensuring that a gold source's own family never supplies a distractor (Alg.~\ref{alg:braid}). A dataset whose items provide no input of their own draws its distractors from charts. The draw is randomly seeded and prefers the least-used sources of the split, so no source dominates the benchmark and the same composition is reproduced on every rebuild.

\paragraph{Quality control.}
A source counts as a \textit{distractor} only if it answers no question of the task: when any other provided source also answers a question, that question has more than one gold source, and thus the task can no longer isolate the \textit{routing} it is built to evaluate. \braid therefore rejects such tasks while composing them (Alg.~\ref{alg:braid}) rather than filtering them after construction, and a task that fails any of the four criteria below is never written. (1) \textit{Separation} keeps no provided source standing in for another: two questions of one task do not rest on near-duplicate gold sources, and a distractor resembling any of the task's gold sources is redrawn. (2) \textit{Routing} requires each question to mention at least one word or phrase that appears in its own gold source and in no other provided source, so that the question can identify that source among those provided. (3) \textit{Leakage} forbids any source other than a question's gold source from containing that question's answer. (4) \textit{Answerability} covers what text matching cannot decide: where a task supplies images or documents, an LLM judges every pairing of a question with a provided source, and the task is discarded if any question's answer holds against a source other than its gold source. Beyond these four criteria, file names are also controlled for the same reason: the provided sources are listed in random order with file names revealing nothing, so a solver can identify the gold source only from its content, never from its file name.

\input{algorithms/braid_construction}

\paragraph{Dataset coverage.}
\braid is constructed upon ten public datasets spanning eight domains \citep{ma2024mmlongbenchdoc,deng2025longdocurl,tanaka2023slidevqa,reddy2024docfinqa,wang2024charxiv,tang2025chartmuseum,hendrycks2021math,he2024olympiadbench,phan2025hle,li2023taco}, including documents, slides, finance, charts, mathematics, physics, science, and code. A dataset enters \braid with whichever \textit{composition methods} and \textit{complexity levels} it supports, and each pairing of a \braid dataset with a composition method forms one \textit{arm}, evaluated at both complexity levels. \textit{Distributed} demands only that each question have its own gold source, which every original dataset supplies, while \textit{anchored} requires a source that carries at least $k$ questions, which the document and slide datasets provide and the others cannot: a dataset whose items are one question per figure, one problem per statement, or a prompt with no provided input has no gold source to share.
A complexity level is kept only where it yields at least $80$ questions. For example, DocFinQA at $k=5$ is dropped for size: each of its filings carries too few questions to compose tasks at that complexity level in sufficient number, giving $5$ questions under \textit{distributed} and $40$ under \textit{anchored}, both far below $80$. Every other absence is a matter of constructibility, since the chart, mathematics, physics, science, and code datasets hold at most a handful of questions per source, and so cannot anchor $k$ of them. What remains spreads difficulty across two axes. Across datasets, the provided sources range from documents and slides to charts and text-only samples, and the answer types from numeric through open-ended to code generation. Within a dataset, raising the complexity level from $k=3$ to $k=5$ carries the same source data into longer tasks, and under \textit{distributed} from five provided sources to nine.

\paragraph{Data statistics.}
\input{tables/appendix_braid_statistics}
As summarized in Tab.~\ref{tab:appendix:braid_statistics}, \braid constructs nineteen arms over fourteen \braid datasets, adapted from ten public datasets, across eight domains. \braid builds workflow-level samples at two complexity levels, counted in questions rather than tasks because accuracy is measured over questions. Every arm is evaluation-only and inherits the test side of the dataset it is composed from, so no \braid task shares a row with any training split.

\subsection{Evaluation Metrics}
\label{appendix:subsec:braid:eval_metrics}

\paragraph{Task Accuracy.}
The accuracy of a task $Q$ is the average score of its $k$ questions, and the accuracy of a dataset $\mathcal{Q}$ is the average over its tasks. Let $\hat y_i$ denote the predicted answer for $q_i$, $y_i$ denote its gold answer, and $t_i$ denote its answer type ($t_i \in \{ \textit{numeric},\textit{math},\textit{text},\textit{choice},\textit{list},\textit{code} \}$):
\begin{equation}
\mathrm{Acc}(Q)=\frac{1}{k}\sum_{i=1}^{k}\mathrm{score}_{t_i}\big(\hat y_i,\,y_i\big),
\qquad
\mathrm{Acc}(\mathcal{Q})=\frac{1}{\lvert\mathcal{Q}\rvert}\sum_{Q\in\mathcal{Q}}\mathrm{Acc}(Q)
\label{eq:braid_accuracy}
\end{equation}

\paragraph{Numeric Answers.}
A numeric answer is scored on its value rather than on its string, and a dataset whose golds are measured quantities may admit a relative tolerance $\varepsilon$:
\begin{equation}
\mathrm{score}_{\textit{numeric}}\big(\hat y,y\big)=\one\big[\lvert\hat y-y\rvert\le\varepsilon\lvert y\rvert\big]
\label{eq:braid_numeric}
\end{equation}

\paragraph{Mathematical Answers.}
A mathematical answer is scored under symbolic equivalence $\equiv$, so an answer differing from the gold in notation, ordering, or simplification still counts:
\begin{equation}
\mathrm{score}_{\textit{math}}\big(\hat y,y\big)=\one\big[\hat y\equiv y\big]
\label{eq:braid_math}
\end{equation}

\paragraph{Textual Answers.}
A textual answer counts in full when it matches the gold after normalization $\mathrm{norm}(\cdot)$, and otherwise earns the token-level $\mathrm{F}_1$ between the two. Thus, an answer carrying part of the gold is credited in part:
\begin{equation}
\mathrm{score}_{\textit{text}}\big(\hat y,y\big)=
\begin{cases}
1 & \text{if } \mathrm{norm}(\hat y)=\mathrm{norm}(y)\\[2pt]
\mathrm{F}_1\big(\hat y,y\big) & \text{otherwise}
\end{cases}
\label{eq:braid_text}
\end{equation}
A dataset whose golds are short spans of a long source may admit the relaxed form, in which a gold appearing inside the answer also counts in full, and where a dataset accepts several golds the answer takes its best score among them.

\paragraph{Multiple-Choice Answers.}
A multiple-choice answer is scored on the option it selects, which is extracted from the answer before comparison:
\begin{equation}
\mathrm{score}_{\textit{choice}}\big(\hat y,y\big)=\one\big[\hat y=y\big]
\label{eq:braid_choice}
\end{equation}

\paragraph{List Answers.}
A list answer is scored position by position against the gold sequence, so an answer right in part is credited in part:
\begin{equation}
\mathrm{score}_{\textit{list}}\big(\hat y,y\big)=\frac{1}{\lvert y\rvert}\sum_{j=1}^{\lvert y\rvert}\one\big[\hat y_j=y_j\big]
\label{eq:braid_list}
\end{equation}

\paragraph{Coding Answers.}
A coding answer is scored by execution against the tests $\mathcal{V}$, and counts only when every test passes:
\begin{equation}
\mathrm{score}_{\textit{code}}\big(\hat y,y\big)=\one\big[\mathrm{pass}(\hat y,v)\ \ \forall v\in\mathcal{V}\big]
\label{eq:braid_coding}
\end{equation}

\paragraph{Per-Question Grading.}
Every question of a task is graded separately according to its answer type, so the accuracy of a task rises with each question whose routing succeeds. Two answer types are graded by degree within a single question as well: a textual answer that does not match its gold falls back to token overlap (Eq.~\ref{eq:braid_text}), and a list answer is credited position by position (Eq.~\ref{eq:braid_list}). Particularly, we grade per question rather than scoring a task as a single unit. This is because a single-unit score counts only when all $k$ of its questions are correct, which therefore assigns the same value to a solver that answers most questions of a task and to one that answers none. Such a score fails to effectively measure the actual performance and competence of a solver in workflow-level problem-solving.




\section{Beneath the Flow: Complementary Details and Discussions}
\label{appendix:sec:inflowop}

\subsection{\ours Overview: From the Flow as Built to the Flow as Run}
\label{appendix:subsec:inflowop:stages}

\ours performs a task in two stages (Alg.~\ref{alg:overview}): \textit{build} the workflow and \textit{run} the workflow. The first \textit{workflow construction stage} turns the task into a workflow: it prices every atom against every agent into the cost matrix, leveraging the cost matrix to determine how finely the task is divided and who carries each divided subtask, and hands every subtask the contract its atoms declare. The \textit{in-flow optimization stage} runs that workflow and corrects it as it runs: it matches each realized subtask against its contract by reusing the estimator in the first stage, blames whichever side breaches, and tempers the workflow there locally. The two stages share one estimator and one matrix, which is what lets a correction be priced in the same currency the workflow was built with.

\paragraph{Stage I: Construction (Alg.~\ref{alg:coalesce}).}
The task is first decomposed to its finest atoms, and the atoms the goal never needs are discarded. \coalesce proceeds over partial groupings, extending each by its next atom, which either opens a new subtask or joins one already open. A grouping is pruned once its lower bound reaches the best cost found so far. Each subtask of the returned atom grouping is assigned with its lowest-cost agent. When no agent meets the competence bar, a new agent is created at the creation penalty. The atom dependencies then carry over to the subtask graph, and every subtask leaves this stage with its contract. This then supports Stage II for label-free in-flow optimization.

\paragraph{Stage II: In-Flow Optimization (Alg.~\ref{alg:inflowop}).}
Execution proceeds subtask by subtask, and every realized input and output is retained. A subtask that yields no usable output is faulty by observation. Otherwise, its realized work is matched against its contract, and which side breaches blames the fault. A fault blamed on the assignment is corrected by the next agent the cost matrix ranks, and one blamed on the decomposition by re-decomposing the subtask under the same contract (Eq.~\ref{eq:ladder}). A correction is kept only if the subtask then executes and its unmet conditions are satisfied, and is undone otherwise. Tempering therefore never leaves a workflow worse than the one Stage I produced. Tempter is confined to the the local subtask (Eq.~\ref{eq:closure}) with every progress achieved so far intact.

\paragraph{Cost-Driven Pruning.}
\coalesce expands a frontier $\mathcal{F}$ of partial coalescings and retains the best complete one found, the incumbent $P^{\star}$. Since $\mathrm{Cost}$ (Eq.~\ref{eq:workflow_cost}) only accumulates as atoms are placed, any completion of a partial coalescing $P'$ is bounded from below by the lower bound $\mathrm{lb}=\mathrm{Cost}(P')+\textsc{OptimisticRemainder}(P')$, where the remainder is the cheapest cost the unplaced atoms could possibly incur. As that remainder never over-estimates, a partial coalescing whose $\mathrm{lb}$ already reaches $\mathrm{Cost}(P^{\star})$ can complete no better and is pruned, without risk of discarding the optimum.

\input{algorithms/A1A2_coalesce}
\input{algorithms/A3A4_inflowop}

\subsection{Meta Primitives}
\label{appendix:subsec:inflowop:meta_primitives}

\coalesce may emit four structures. Throughout, $\hat p(s,a)$ is the probability that agent $a$ realizes subtask $s$ (\S\ref{subsec:method:decomposition}), $C(s,a)=-\log\hat p(s,a)$, and $\ell(s,a)$ its latency.

\paragraph{\textsc{And}} for \textit{conjunction}: atoms are grouped into one subtask, and subtasks with no dependency path between them are left unordered. Its reliability and latency are Eq.~\ref{eq:reliability} and Eq.~\ref{eq:latency} unchanged.

\paragraph{\textsc{Repeat}} for \textit{bounded repetition}: a subtask is re-run in place until its output is usable, at most $B_{\mathrm{rep}}$ times.
\begin{equation}
R_{\textsc{rep}}=-\log\big(1-(1-\hat p)^{B_{\mathrm{rep}}}\big),\qquad
L_{\textsc{rep}}=\ell\sum_{r=1}^{B_{\mathrm{rep}}}(1-\hat p)^{\,r-1}
\label{eq:mp_repeat}
\end{equation}

\paragraph{\textsc{Or}} for \textit{conditional alternatives}: one required output is pursued by $J$ independent methods and consolidated.
\begin{equation}
R_{\textsc{or}}=-\log\Big(1-\prod_{j=1}^{J}\big(1-\hat p_j\big)\Big),\qquad
L_{\textsc{or}}=\max_{j\le J} L_j
\label{eq:mp_or}
\end{equation}

\paragraph{\textsc{Select}} for \textit{branched selection}: exactly one of $K$ mutually exclusive continuations is taken under a branch test.
\begin{equation}
R_{\textsc{sel}}=\sum_{j=1}^{K} \omega_j R_j,\qquad
L_{\textsc{sel}}=\max_{j\le K} L_j
\label{eq:mp_select}
\end{equation}
where $R_j$ and $L_j$ are the reliability and latency of continuation $j$, and $\omega_j$ its selection weight.

\subsection{Taxonomy of Cost}
\label{appendix:subsec:inflowop:cost_taxonomy}

This section complements our cost taxonomy definition in \S\ref{sec:method}. A constructed workflow is charged along two axes: the \textit{reliability cost} $R$ of entrusting each subtask to an agent that may not realize it correctly, and the \textit{latency cost} $L$ of running the agents it assigns. Both follow from the same label-free matching of what a subtask demands against what an agent supplies (\S\ref{subsec:method:decomposition}), so a workflow is priced before it runs. Together with the price of each agent created for it, they compose the \textit{workflow cost} $\mathrm{Cost}(W)$ that \coalesce minimizes (Eq.~\ref{eq:construction_objective}). The \textit{optimization cost} of the taxonomy stays outside $\mathrm{Cost}(W)$, as what improving a workflow charges rather than what running one charges (\S\ref{appendix:sec:implementation}).

\paragraph{Reliability cost.}
A workflow succeeds only when its subtasks are each realized correctly. We model the realization of a subtask $s_i$ by its assigned agent $a(s_i)$ as a success with probability $\hat p\big(s_i,a(s_i)\big)\in(0,1]$. As subtask realizations are independent, a workflow's success probability factorizes as $\prod_{s_i}\hat p(s_i,a(s_i))$. Thus, maximizing it is equivalent to minimizing its negative logarithm:
\begin{equation}
R(W)=-\log\!\prod_{s_i\in S}\hat p\big(s_i,a(s_i)\big)=\sum_{s_i\in S} C\big(s_i,a(s_i)\big),
\qquad C(s,a):=-\log\hat p(s,a)
\label{eq:reliability}
\end{equation}
where $C(s,a)$ is the \emph{reliability cost} of realizing $s$ with $a$, and $R(W)$ is the workflow's total reliability cost.

\paragraph{Latency cost.}
Each subtask also incurs a latency $\ell\big(s_i,a(s_i)\big)$, the cost of running its agent. Because $G$ admits parallelism, a workflow's latency is not the sum of these but its \textit{critical path} (\S\ref{subsec:method:decomposition}):
\begin{equation}
L(W)=\max_{\text{path }\pi\subseteq G}\;\sum_{s_i\in\pi}\ell\big(s_i,a(s_i)\big)
\label{eq:latency}
\end{equation}
Parallel subtasks thus cost no more than the slowest among them, so spreading work across concurrent agents lowers $L$ even where it raises $R$.

\paragraph{Workflow cost.}
The cost of a workflow combines the two axes, reliability (Eq.~\ref{eq:reliability}) and latency (Eq.~\ref{eq:latency}), scalarized by a trade-off weight $\beta$, with a penalty $\gamma$ for each newly created agent (\S\ref{subsec:prelim:build}):
\begin{equation}
\mathrm{Cost}(W)=\underbrace{R(W)}_{\text{reliability}}+\;\beta\underbrace{L(W)}_{\text{latency}}
+\;\gamma\,\lvert\mathcal{A}_{+}(W)\rvert
\label{eq:workflow_cost}
\end{equation}
where $\mathcal{A}_+(W)$ is the set of agents created for $W$, $\beta$ trades reliability against latency, and $\gamma$ prices each creation. Sweeping $\beta$ traces a reliability-latency Pareto frontier: small $\beta$ favors reliable, finer-decomposed workflows; while large $\beta$ favors fast, compact ones.

\subsection{A Step Is Worth What It Leads To}
\label{appendix:subsec:inflowop:outcome_only}

\input{figures/appendix_judge_by_outcome}

\begin{hlbox}{{The Verdict Lies Downstream, Not in the Middle}}
\textit{Whether an intermediate output helps or harms cannot be determined from that output alone.} A step may satisfy the conditions it is expected to meet yet still steer the flow wrong, while it may also violate them yet still bridge the flow to the correct final answer. The sign of a deviation is thus settled by where it drives the rest of the flow, not by the step where it appears or by what that step produces. Therefore, no local judgment of that step, however it is computed, recovers it.
\end{hlbox}
\vspace{6pt}

Figure~\ref{fig:appendix:judge_by_outcome} shows one such case: the subtask asks which images carry a particular set of information, so its declared condition is that the output identify the useful ones.
Before tempering, the agent names two images with wrong evidence but successfully meeting that condition. Thus, neither image carries the information the subtask demands, and the flow continues on wrong evidence that misleads its final answer. After tempering, the agent responds that no image contains what is required. That answer violates the condition, yet it is the accurate one. Consequently, the flow carries on with it and reaches the correct final answer. \textit{However, if scored against its declared condition, the step that helped is recorded as the step that failed}.

\textit{We therefore judge every temper by the outcome of the flow it belongs to, without relying on step-level assessment.} A step-level metric is not a reliable or stricter measure of the same quantity, as on cases of this kind it carries the opposite sign. For the same reason, we do not evaluate on benchmarks that score intermediate steps against annotated ground truth \citep{chen2026seeingelephantbenchmarkfailure}, since such a benchmark penalizes exactly the behavior this example shows to be correct.

\subsection{Temper Locally, Look Globally}
\label{appendix:subsec:inflowop:global}

Local detection (\S\ref{subsec:method:inflowop}) acts where a fault breaches the contract of its own subtask. However, a workflow may nonetheless terminate without a usable answer for a cause lying several steps upstream, without violating local contract. Rather than fall back on a global re-search, we extend in-flow optimization with an optional \emph{global localization}: the \textit{temper} stays local, and only \emph{where to look} becomes global. Once the local pass ends without a usable answer, the whole graph is scanned to rank subtasks by \emph{suspicion}:
\begin{equation}
\mathrm{susp}(s_i)=\big(1-\hat p_i\big)\big(1+\lvert\mathrm{Desc}(s_i)\rvert\big)+\one\!\left[s_i\ \text{flagged}\right]
\label{eq:suspicion}
\end{equation}
where a subtask already flagged by local detection receives an additive bonus. Suspicion prefers subtasks both unreliable and consequential: an unreliable subtask whose output many others consume is a likelier root cause than an unreliable leaf. The most suspicious untried subtask is then handed to the cost-ordered ladder (Eq.~\ref{eq:ladder}), committed under the same evidence, and followed by a scoped resumption of its closure (Eq.~\ref{eq:closure}), until a usable answer emerges or the probe budget $B_{\mathrm{g}}$ is spent (Alg.~\ref{alg:global}). Global localization therefore buys only \emph{where to look} at the final failure, while every correction is committed locally. It is disabled in all main experiments, and we evaluate it as an ablation (\S\ref{subsec:exps:ablation}).

\input{algorithms/global_localization}

\section{Implementation Details}
\label{appendix:sec:implementation}

\textbf{Matched Compute Budget.}
The compute budget each single agent baseline receives is matched to the workflow it is compared against on every task rather than on average, so no task lets the workflow spend what the baseline could not.

\textbf{Configuration.}
We summarize our experiment configurations in Tab.~\ref{tab:appendix:configuration}, including the core parameters' definitions and configurations, as well as the default values our ablation studies explore against.

\textbf{Cost Accounting.}
For any system, its problem-solving cost of a task is the total number of tokens the system consumes, including both input and output over construction, execution, and in-flow optimization.

\textbf{Accuracy Against Cost.}
A commonly used approach for improving workflow performance is to best-of-$N$, optimizing workflows by rerunning workflow construction and execution $N$ times and and keeping the best outcome. However, this optimization gains limited accuracy at the cost of $\sim N \times$ compute cost. To optimize cheaply, we propose \ours that optimizes locally in flow without global reruns of any construction or execution stage. To compare the effectiveness of different methods, we study on the single-agent baseline, best-of-$N$, and \ours at the matched compute cost by giving the same turn budget. Fig.~\ref{fig:exp_cost} compares best-of-$N$ ($N=4$) and \ours against both the single agent baseline they extend and the single LLM baseline showing the backbone competence, with every cost stated as a multiple of one single-agent run on the same backbone. At matched turn budget, \ours gains $+7.15\%$, $+7.75\%$, $+10.68\%$, $+11.97\%$, and $+7.38\%$ over the single agent on \texttt{Qwen3.5-4B}, \texttt{Qwen3.5-9B}, \texttt{GPT-5-mini}, \texttt{GPT-5.4-mini}, and \texttt{GPT-5.6-luna}, respectively. Best-of-$N$ spends $4\times$ those turns and reaches $15.83\%$, $20.63\%$, $21.72\%$, $21.25\%$, and $43.82\%$, consistently staying below \ours on every backbone.
This indicates that a gain follows from how a workflow is constructed and optimized rather than from how many times it is attempted.

\input{figures/exp_cost}

\input{tables/appendix_configuration}

%% file: sections/6_results.tex




%% file: sections/7_related_work.tex
\section{Related Work}
\label{sec:related_work}

\subsection{Workflow Construction over an Agent Pool}
\label{subsec:related_work:construction}

Compared to hand-designed multi-agent systems that fix the roles and their order in advance \citep{hong2024metagpt,li2026agentflow,guo2026anagent}, automatic construction searches for the effective structure rather than authoring it, within a design space of operators, connections, or collaboration structures enumerated in advance \citep{zhang2025aflow,zhao2025a2flow,yue2026autoras,li2025argdesigner}. That design space, rather than the agent pool, settles how finely the task is divided: prompted decomposition takes its grain from exemplars or instructions \citep{liu2025selectthendecompose,madhwal-2026-decomposed-prompting,wang2023planandsolveprompting}, while plan-then-assign pipelines fix the parts before choosing which agent runs each one \citep{shen2023hugginggpt,dong2026hybridflow,kapoor2026trim}.
Where the grain adapts, it adapts to the query alone \citep{su2026daao}, or revises a division already proposed \citep{prasad2024adapt,wang2025tdag}.
Cost enters some of these decisions, in recent work alongside how far the task is divided \citep{zhang2025maas,cui2026unoorchestra,xu2026tipex}, but \textit{the quantity that prices a choice is fitted before the task and refitted whenever the set of available operators changes}.

Which agent carries out each part divides along a second axis.
Some methods search over a pool they leave fixed, so a part that no agent fits well still goes to the closest one available \citep{zhuge2024gptswarm,yuan2025knapsack}. Another line of work adds agents that the pool does not hold, so the same part instead yields an agent created for it \citep{zhang2025swarmagentic,wu2025kamac}.
Where a system offers both, it alternates between them or settles the choice by a test of similarity, rather than weighing one against the other \citep{shang2025agentsquare,gupta2026fasta,zhang2026agenticplancaching}. As for pricing, where it appears, it ranges over backbone models rather than over reuse and creation \citep{yang2025bamas}.
\textit{What is missing is a single price over the pool as it stands, charged per part before execution, under which creating an agent is one option inside the same search that sets the grain}.

\subsection{Optimizing a Workflow after Execution}
\label{subsec:related_work:optimization}

Recent work optimizes workflow performance in different ways.
One route improves a workflow by training: the constructor or planner is fine-tuned against scored outcomes \citep{li2026agentflow,peng2026hiper,nielsen2026conductor,dang2025puppeteer}. Step-level signals are obtained the same way, with process reward models and learned verifiers fitted on annotated steps or on rollouts graded against known answers \citep{zhang2026webarbiter,chae2025webshepherd,zhu2026factutility,lee2026prints}.
\textit{Improvement is therefore paid for before the task arrives, in data and compute}.
A second route requires no training but re-runs instead, sampling the workflow repeatedly, searching for a better one, or re-executing after a failure \citep{hwang2026agenticmontecarlosimulating,li2025encompass,yoon2026patplanningaftertrialefficienttesttime,lee2026aggagent}.
\textit{Each increment of quality costs another execution of the entire workflow, while what needs correcting is only a small part of it}.
A third route is both training-free and label-free: the system criticizes its own output and revises it \citep{shi2026vlmsseeingjustsaying,yang2025theory4selfcorrection,ning2026revisionresolvingdecomposingsecondpass,wan2026deepverifier}.
\textit{Yet what it revises is the flow output, without identifying which part inside the flow is responsible}.
When blame is assigned within a flow, it is scored on a completed trace against annotated ground truth \citep{chen2026seeingelephantbenchmarkfailure}, recovered by replaying a trace that has already finished \citep{hou2026edgeerrordependencygraphguided,lin2026reflectinterventionsupportederrorattribution}, or derived from a relation the domain itself makes checkable, such as whether a claim follows from the sources cited for it \citep{hirsch2026agentblamelocalizingfaithfulness}.
\textit{What is missing is a fault located from a single execution as it runs, with the correction confined to the part at fault, under no training, no reference answer, and no graded outcome}.

\subsection{Evaluating Multi-Agent Workflows}
\label{subsec:related_work:evaluation}

Workflow systems are measured on the suites built for single models: mathematical reasoning, code generation, and question answering \citep{zhang2025aflow,zhang2026radar,yun2026graphofagents,zhou2026mass}.
On these, repeated sampling from one strong agent captures much of what coordination adds, and under matched budgets a coordinating team can fall behind the same models working alone \citep{ann2026interactiontaxcommunicationerases,sun2026collaborationtaxllmmultiagent,choi2025debateorvote,jwalapuram2026illusionmultiagentadvantage}.
\textit{A benchmark that one agent already solves cannot show what dividing the work contributes}.

Benchmarks that do stress coordination exist, built around long-horizon tool use, siloed state, or distributed evidence \citep{gao2026swarmbench,zhang2026silobench,chen2026appworld}.
In these, the need to divide is imposed by the environment rather than arising from what the task demands: state is partitioned, tools are split between agents, or knowledge is withheld from one of them.
\textit{What is missing is a benchmark whose tasks require the division on their own terms, with parts that call for different competences and a result that depends on how those parts combine}.

%% file: figures/appendix_braid_necessity.tex
\begin{figure}[H]
    \vspace{20pt}
    
    \small
    \centering

    \includegraphics[width=\textwidth]{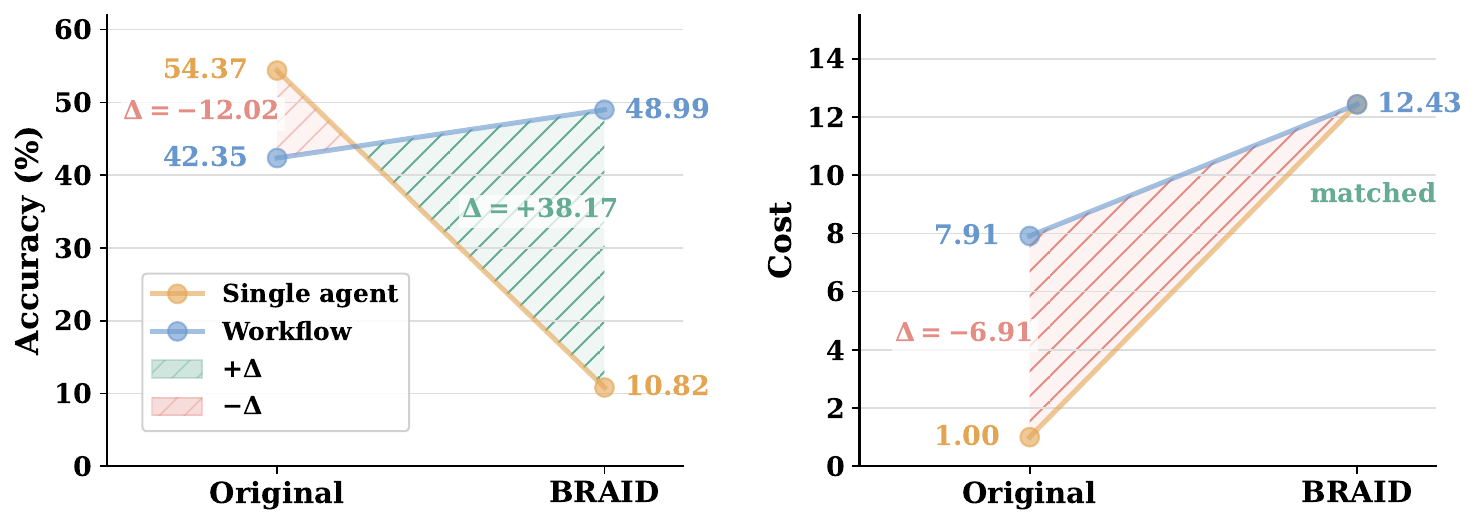}

    \caption{\textbf{A Workflow Earns Its Necessity Only Where One Agent Falls Short.} The shaded band marks the gap between the two solvers in accuracy (left) and in cost (right). Cost is counted in units of one single-agent run on the original data, and on \braid the two solvers run under matched budgets. Both use \texttt{gpt-5-mini} over 150 samples of \textit{SlideVQA} (\textit{doc} variant of \braid), \textit{Distributed} at complexity level $k=5$.}
    \label{fig:appendix:braid_necessity}

    \vspace{16pt}
\end{figure}

%% file: algorithms/braid_construction.tex
\begin{algorithm}[t]

\caption{\textsc{Compose}$(\mathcal{S},k)$: \braid Task Construction (\S\ref{appendix:sec:braid})}
\label{alg:braid}

\begin{algorithmic}[1]
\Require the questions and sources $\mathcal{S}$ of one dataset split; complexity level $k$; composition \textit{distributed} or \textit{anchored}
\Statex \textcolor{commentcolor}{\textit{\textbf{Notation:} $\mathrm{gold}(q_i)$ is the source $q_i$ is answered from, $y_i$ is its answer, $w\in q_i$ is a word or phrase of its text, $\mathrm{sim}(\cdot,\cdot)$ is source similarity with bar $\theta$, and $\mathrm{judge}(q_i,z)\in\{0,1\}$ is an LLM judge on whether $y_i$ holds against $z$}}
\Ensure task $Q=\langle\{q_1,\dots,q_k\},\,E\cup N\rangle$, or \textsc{Discard}

\State $\{q_1,\dots,q_k\}\gets$ $k$ questions of the split \textbf{ s.t. } $\lvert\{\mathrm{gold}(q_i)\}\rvert=k$ (\textit{distributed}) \textbf{ or } $1$ (\textit{anchored})
\State $E\gets\{\mathrm{gold}(q_i)\}$;\quad $N\gets\textsc{Draw}\big(\mathcal{S}\setminus E,\;k-1\big)$ \Comment{\textit{gold sources}; \textit{distractors}, drawn seeded and least-used first}

\Statex \textcolor{commentcolor}{\textit{(1) \textbf{Separation}: no provided source stands in for another}}
\If{$\exists\,e\neq e'\in E:\ \mathrm{sim}(e,e')\ge\theta$}
   \State \Return \textsc{Discard} \Comment{one question's gold source would answer another}
\EndIf
\ForAll{$d\in N$ \textbf{ with } $\max_{e\in E}\mathrm{sim}(d,e)\ge\theta$}
   \State $d\gets\textsc{Draw}\big(\mathcal{S}\setminus(E\cup N),\,1\big)$, \textbf{ or } \Return \textsc{Discard} \Comment{redraw until clean, else drop}
\EndFor

\Statex \textcolor{commentcolor}{\textit{(2)-(4) \textbf{Answerability}: exactly one provided source answers each question}}
\ForAll{$q_i$ \textbf{ with } $Z_i\gets(E\cup N)\setminus\{\mathrm{gold}(q_i)\}$}
   \If{$\nexists\,w\in q_i:\ w\in\mathrm{gold}(q_i)\ \wedge\ w\notin z\ \ \forall z\in Z_i$}
      \State \Return \textsc{Discard} \Comment{(2) \textit{routing}: no word identifies $\mathrm{gold}(q_i)$ among $E\cup N$}
   \EndIf
   \If{$\exists\,z\in Z_i:\ y_i\subseteq z$}
      \State redraw $z$ \textbf{ if } $z\in N$, \textbf{ else } \Return \textsc{Discard} \Comment{(3) \textit{leakage}}
   \EndIf
   \If{$E\cup N$ carries images or documents \textbf{ and } $\exists\,z\in Z_i:\ \mathrm{judge}(q_i,z)=1$}
      \State \Return \textsc{Discard} \Comment{(4) \textit{answerability}, where $y_i\subseteq z$ cannot be decided by text}
   \EndIf
\EndFor

\State \Return $Q\gets\langle\{q_1,\dots,q_k\},\ \textsc{Shuffle}(E\cup N)\rangle$ \Comment{random order, file names carrying no signal}
\end{algorithmic}

\end{algorithm}

%% file: tables/appendix_braid_statistics.tex
\begin{table}[t]
\vspace{-32pt}

\centering
\small
\caption{\textbf{Data Statistics.} \braid as built: $19$ evaluation-only arms over $8$ domains, with questions each yields at two complexity levels $k=3$ and $k=5$. \textit{Original Dataset} shows where a \braid task is adapted from. For SlideVQA, we construct two variants: (1) \textit{SlideVQA-Doc}: providing slides as a document, and (2) \textit{SlideVQA-Img}: providing slides as individual images. \textit{Task Type} is the answer type a \braid sample's $k$ questions demand, and \textit{Data Type} is the type of sources provided beside input questions. A level is absent where the dataset yields too few questions ($n_q<80$) to evaluate on, and \textit{Anchored} is absent where no source of that dataset carries $k$ questions to anchor.}
\label{tab:appendix:braid_statistics}
\vspace{2pt}

\begin{tabularx}{\textwidth}{@{}XXccccc@{}}
\toprule
\multicolumn{1}{c}{\textbf{Original Dataset}} & \multicolumn{1}{c}{\textbf{\braid}} & \textbf{Composition} & \textbf{Task Type} & \textbf{Data Type} & \multicolumn{1}{c}{$\bm{k=3}$} & \multicolumn{1}{c}{$\bm{k=5}$} \\
\midrule
\noalign{\vskip 2pt}
\rowcolor{blue_bg}\multicolumn{7}{c}{\textbf{\textit{\textcolor{blue_text}{Document}}}} \\
\noalign{\vskip 2pt}
MMLongBench-Doc~\citep{ma2024mmlongbenchdoc} & MMLongBench-Doc & Distributed & Mixed & Document & 426 & 175 \\
MMLongBench-Doc~\citep{ma2024mmlongbenchdoc} & MMLongBench-Doc & Anchored & Mixed & Document & 498 & 305 \\
LongDocURL~\citep{deng2025longdocurl} & LongDocURL & Distributed & Mixed & Document & 441 & 340 \\
LongDocURL~\citep{deng2025longdocurl} & LongDocURL & Anchored & Mixed & Document & 399 & 325 \\
\noalign{\vskip 2pt}
\rowcolor{blue_bg}\multicolumn{7}{c}{\textbf{\textit{\textcolor{blue_text}{Slide}}}} \\
\noalign{\vskip 2pt}
SlideVQA~\citep{tanaka2023slidevqa} & SlideVQA-Doc & Distributed & Open-Ended & Document & 1557 & 1000 \\
SlideVQA~\citep{tanaka2023slidevqa} & SlideVQA-Doc & Anchored & Open-Ended & Document & 1470 & 970 \\
SlideVQA~\citep{tanaka2023slidevqa} & SlideVQA-Img & Distributed & Open-Ended & Image & 1581 & 1005 \\
SlideVQA~\citep{tanaka2023slidevqa} & SlideVQA-Img & Anchored & Open-Ended & Image & 1470 & 1000 \\
\noalign{\vskip 2pt}
\rowcolor{blue_bg}\multicolumn{7}{c}{\textbf{\textit{\textcolor{blue_text}{Finance}}}} \\
\noalign{\vskip 2pt}
DocFinQA~\citep{reddy2024docfinqa} & DocFinQA & Distributed & Numeric & Document & 84 & -- \\
DocFinQA~\citep{reddy2024docfinqa} & DocFinQA & Anchored & Numeric & Document & 192 & -- \\
\noalign{\vskip 2pt}
\rowcolor{blue_bg}\multicolumn{7}{c}{\textbf{\textit{\textcolor{blue_text}{Chart}}}} \\
\noalign{\vskip 2pt}
CharXiv~\citep{wang2024charxiv} & CharXiv & Distributed & Mixed & Image & 429 & 430 \\
ChartMuseum~\citep{tang2025chartmuseum} & ChartMuseum & Distributed & Mixed & Image & 807 & 805 \\
\noalign{\vskip 2pt}
\rowcolor{blue_bg}\multicolumn{7}{c}{\textbf{\textit{\textcolor{blue_text}{Math}}}} \\
\noalign{\vskip 2pt}
MATH~\citep{hendrycks2021math} & MATH (Level 5) & Distributed & Numeric & Text & 603 & 605 \\
OlympiadBench~\citep{he2024olympiadbench} & OlympiadBench-Math (Text) & Distributed & Numeric & Text & 660 & 660 \\
OlympiadBench~\citep{he2024olympiadbench} & OlympiadBench-Math (Multimodal) & Distributed & Numeric & Image & 150 & 150 \\
\noalign{\vskip 2pt}
\rowcolor{blue_bg}\multicolumn{7}{c}{\textbf{\textit{\textcolor{blue_text}{Physics}}}} \\
\noalign{\vskip 2pt}
OlympiadBench~\citep{he2024olympiadbench} & OlympiadBench-Physics (Text) & Distributed & Numeric & Text & 234 & 200 \\
OlympiadBench~\citep{he2024olympiadbench} & OlympiadBench-Physics (Multimodal) & Distributed & Numeric & Image & 456 & 455 \\
\noalign{\vskip 2pt}
\rowcolor{blue_bg}\multicolumn{7}{c}{\textbf{\textit{\textcolor{blue_text}{Science}}}} \\
\noalign{\vskip 2pt}
HLE~\citep{phan2025hle} & HLE & Distributed & Mixed & Text & 1470 & 785 \\
\noalign{\vskip 2pt}
\rowcolor{blue_bg}\multicolumn{7}{c}{\textbf{\textit{\textcolor{blue_text}{Coding}}}} \\
\noalign{\vskip 2pt}
TACO~\citep{li2023taco} & TACO & Distributed & Code Generation & Text & 1362 & 1360 \\
\bottomrule
\end{tabularx}
\end{table}

%% file: algorithms/A1A2_coalesce.tex
\begin{algorithm}[t]

\caption{\textsc{Coalesce}$(Q,\mathcal{A})$: Bidirectional Cost-Driven Atomic Coalescing (\S\ref{subsec:method:decomposition})}
\label{alg:coalesce}

\begin{algorithmic}[1]
\Require task $Q$, agent pool $\mathcal{A}$, trade-off $\beta$, creation penalty $\gamma$, competence bar $\bar C$, expansion budget $B$
\Ensure workflow $W=\langle S,A,G\rangle$ of least cost (Eq.~\ref{eq:construction_objective})

\Statex \textcolor{commentcolor}{\textit{\textbf{Direction 1:} top-down atomic decomposition}}
\State $\mathcal{T},\,G_{\mathcal{T}}\gets\textsc{Atomize}(Q)$ \Comment{atoms $\{\tau_1,\dots,\tau_n\}$ and their data-flow DAG}
\State $\nu\gets\{\tau\in\mathcal{T}\ :\ \tau\ \text{back-reachable from the goal over}\ G_{\mathcal{T}}\}$ \Comment{drop atoms the goal never needs}

\Statex \textcolor{commentcolor}{\textit{\textbf{Direction 2:} bottom-up cost-driven coalescing}}
\State $C,\ell,\rho\gets\textsc{CostMatrix}(\nu,\mathcal{A})$ \Comment{label-free estimator on singleton atoms}
\State $P^{\star}\gets\{\{\tau\}:\tau\in\nu\}$;\quad $\mathcal{F}\gets\{\varnothing\}$ \Comment{incumbent is the finest cut, one subtask per atom; frontier to extend}
\While{$\mathcal{F}\neq\varnothing$ \textbf{and} explored nodes $<B$}
  \State $P\gets\textsc{Pop}(\mathcal{F})$ \Comment{a partial coalescing; $\tau_i$ is its next atom in topological order}
  \ForAll{$\mu\in\{\textsc{Open}(\tau_i)\}\cup\{\textsc{Merge}(\tau_i,s)\ :\ s\in P\ \text{open}\}$} \Comment{open or merge (\S\ref{subsec:method:decomposition})}
     \State $P'\gets \mu(P)$ \Comment{the resulting coalescing; $s$ is the subtask changed by $\mu$}
     \State $C(s,\cdot),\ell(s,\cdot),\rho(s)\gets\textsc{CostMatrix}(\{s\},\mathcal{A})$ \Comment{re-score only the changed subtask}
     \If{$\rho(s)=1$} \Comment{no pooled agent covers $s$ (Eq.~\ref{eq:coverage})}
        \State $a_{+}\gets\textsc{CreateAgent}(s)$;\quad $\mathcal{A}_{+}\gets\mathcal{A}_{+}\cup\{a_{+}\}$ \Comment{charged $\gamma$ in $\mathrm{Cost}$}
     \EndIf
     \State $a(s)\gets\arg\min_{a\in\mathcal{A}\cup\mathcal{A}_{+}}C(s,a)$ \Comment{bind $s$ to its lowest-cost covering agent}
     \State $\mathrm{lb}\gets\mathrm{Cost}(P')+\textsc{OptimisticRemainder}(P')$ \Comment{lower bound ($\mathrm{lb}$): cost committed so far}
     \If{$\mathrm{lb}<\mathrm{Cost}(P^{\star})$}
        \State $\mathcal{F}\gets\mathcal{F}\cup\{P'\}$;\quad \textbf{if} $P'$ is complete \textbf{then} $P^{\star}\gets P'$
     \EndIf \Comment{$\mathrm{lb}$ never over-estimates, so pruning cannot discard the optimum}
  \EndFor
\EndWhile

\Statex \textcolor{commentcolor}{\textit{Emit}}
\State $S,A\gets$ the subtasks of $P^{\star}$ and their assignments
\State $G\gets\textsc{Lift}(G_{\mathcal{T}},\,S)$ \Comment{lift the atom DAG to the subtask level}
\ForAll{$s_i\in S$}
  \State $\sigma^{i}_{\mathrm{in}},\sigma^{i}_{\mathrm{out}}\gets$ input/output conditions of $s_i$'s member atoms \Comment{contract checked in \S\ref{subsec:method:inflowop}}
\EndFor
\State \Return $W=\langle S,A,G\rangle$

\Statex \textcolor{commentcolor}{\textit{\textbf{Subroutine:} cost matrix}}
\Function{CostMatrix}{$S,\mathcal{A}$}
  \ForAll{$s\in S$ \textbf{and} $a\in\mathcal{A}$}
     \State $\hat p(s,a)\gets\mathrm{match}(s,a)$ \Comment{match the subtask's demands against the agent's card}
     \State $C(s,a)\gets-\log\hat p(s,a)$;\quad $\ell(s,a)\gets$ latency of $a$ on $s$ \Comment{Eq.~\ref{eq:reliability}}
  \EndFor
  \ForAll{$s\in S$}
     \State $\rho(s)\gets\one\big[\min_{a}C(s,a)>\bar C\big]$ \Comment{coverage residual (Eq.~\ref{eq:coverage}): nothing in the pool fits $s$}
  \EndFor
  \State \Return $C,\ell,\rho$
\EndFunction

\end{algorithmic}

\end{algorithm}

%% file: algorithms/A3A4_inflowop.tex
\begin{algorithm}[t]

\caption{\textsc{InFlowOp}$(W,C)$: In-Flow Dynamic Optimization (\S\ref{subsec:method:inflowop})}
\label{alg:inflowop}

\begin{algorithmic}[1]
\Require workflow $W=\langle S,A,G\rangle$ with contracts $\{(\sigma^{i}_{\mathrm{in}},\sigma^{i}_{\mathrm{out}})\}$, cost matrix $C$, unmet bar $\bar u$, re-assign budget $B_{\mathrm{ra}}$, re-decompose $B_{\mathrm{rd}}$
\Ensure final answer, and the tempered workflow $W^{\dagger}$ (Eq.~\ref{eq:inflowop_objective})
\State $\mathrm{Mem}\gets\varnothing$;\quad $\mathcal{X}\gets\varnothing$ \Comment{realized results; breached pairings}

\Statex \textcolor{commentcolor}{\textit{\textbf{Local:} detect and temper in flow}}
\ForAll{$s_i\in S$ in topological order of $G$}
  \If{$s_i\in\mathrm{Mem}$} \Comment{\textsc{Resume}: a valid result is never recomputed}
     \State \textbf{continue}
  \EndIf
  \State $(\mathcal{I}_i,\mathcal{O}_i)\gets\textsc{Execute}(s_i,a(s_i))$;\quad $\mathrm{Mem}[s_i]\gets(\mathcal{I}_i,\mathcal{O}_i)$
  \State $b\gets\textsc{Detect}(s_i,\mathcal{I}_i,\mathcal{O}_i)$
  \If{$b\neq\textit{none}$}
     \State $\textsc{Pause}(s_i)$ \Comment{halt at the fault, not at the flow}
     \State $W\gets\textsc{Temper}(W,s_i,b)$;\quad $\mathrm{Mem}\gets\mathrm{Mem}\setminus I(s_i)$ \Comment{invalidate only the closure (Eq.~\ref{eq:closure})}
  \EndIf
\EndFor
\State \Return final answer;\quad $W^{\dagger}\gets W$

\Statex \textcolor{commentcolor}{\textit{\textbf{Subroutine:} detection}}
\Function{Detect}{$s_i,\mathcal{I}_i,\mathcal{O}_i$}
  \If{$\mathcal{O}_i$ unusable} \Comment{causal signal: faulty by observation, no matching needed}
     \State \Return \textit{assignment}
  \EndIf
  \State $U_{\mathrm{in}}\gets\{\sigma\in\sigma^{i}_{\mathrm{in}}:u(\sigma,\mathcal{I}_i)\ge\bar u\}$;\quad $U_{\mathrm{out}}\gets\{\sigma\in\sigma^{i}_{\mathrm{out}}:u(\sigma,\mathcal{O}_i)\ge\bar u\}$ \Comment{credit matrix (Eq.~\ref{eq:credit})}
  \State \Return $\mathrm{blame}(s_i)$ given by $U_{\mathrm{in}},U_{\mathrm{out}}$ \Comment{Eq.~\ref{eq:fault_type}}
\EndFunction

\Statex \textcolor{commentcolor}{\textit{\textbf{Subroutine:} tempering along the cost-ordered ladder}}
\Function{Temper}{$W,s_i,b$}
  \If{$b=\textit{assignment}$} \Comment{\textsc{Re-assign}: the cheaper rung (Eq.~\ref{eq:ladder})}
     \For{$b=1$ \textbf{to} $B_{\mathrm{ra}}$}
        \State $a'\gets\arg\min_{a\notin\mathcal{X}(s_i)}C(s_i,a)$;\quad $\mathcal{X}(s_i)\gets\mathcal{X}(s_i)\cup\{a'\}$ \Comment{Eq.~\ref{eq:reassign}}
        \State $(\mathcal{I}_i,\mathcal{O}_i)\gets\textsc{Execute}(s_i,a')$
        \If{$\textsc{Detect}(s_i,\mathcal{I}_i,\mathcal{O}_i)=\textit{none}$} \Comment{commit only on a clean pass}
           \State $a(s_i)\gets a'$;\quad \Return $W$
        \EndIf
     \EndFor
     \State restore $a(s_i)$ \Comment{no unverified change survives}
  \EndIf
  \For{$b=1$ \textbf{to} $B_{\mathrm{rd}}$} \Comment{\textsc{Re-decompose}: climb, no agent can satisfy $s_i$ as scoped}
     \State $\langle S_{\mathrm{sub}},G_{\mathrm{sub}}\rangle\gets\textsc{Coalesce}\big(s_i\mid\sigma^{i}_{\mathrm{in}},\sigma^{i}_{\mathrm{out}}\big)$ \Comment{Alg.~\ref{alg:coalesce}, re-cut under the same contract}
     \State $W\gets\textsc{Splice}(W,s_i,\langle S_{\mathrm{sub}},G_{\mathrm{sub}}\rangle)$;\quad \Return $W$ \Comment{the contract keeps the splice local}
  \EndFor
  \State \Return $W$ \Comment{budget spent: keep the original}
\EndFunction

\end{algorithmic}

\end{algorithm}

%% file: figures/appendix_judge_by_outcome.tex

\definecolor{jbook}{HTML}{2E7D32}
\definecolor{jbobad}{HTML}{C62828}

\begin{figure}[t]
\centering
\small
\begin{tcolorbox}[colback=bboxface, colframe=bboxedge, boxrule=0.6pt, arc=2pt, left=6pt, right=6pt, top=5pt, bottom=5pt]

{\bfseries\textcolor{bboxtag}{Subtask}}\quad \textit{declared condition:} the output identifies images that contain required information.\\[2pt]
{\ttfamily\scriptsize SYSTEM}\quad You are the \texttt{image\_analyzer} agent. Your job is to execute the assigned instruction faithfully and \texttt{[...]} \\
{\ttfamily\scriptsize ...} \\
{\ttfamily\scriptsize USER}\quad$\text{ }\text{ }$ Which images contain the "annual net gains" that \texttt{[...]} \textit{\# Inputs:} image~1: \texttt{[...]}, image~2: \texttt{[...]}, \texttt{[...]}.

\vspace{4pt}\hrule\vspace{4pt}

{\bfseries Before tempering}\\
{\ttfamily\scriptsize ASSISTANT}\quad \texttt{[...]} image~2 (\texttt{/workspace/bussiness\_page12\_crop4.png}) plots "gross income" across 12 months; image~5 (\texttt{/workspace/bussiness\_page17\_crop1.png}) plots market share variations across 2024-2025. Other images show no annual information. So only image~2 and image~5 contain the required information. \texttt{[...]}\\[1pt]
\textcolor{jbook}{\ding{51}}~\textit{condition met} \qquad \textcolor{jbobad}{\ding{55}}~\textit{final answer wrong}

\vspace{4pt}\hrule\vspace{4pt}

{\bfseries After tempering}\\
{\ttfamily\scriptsize ASSISTANT}\quad \texttt{[...]} No image can provide the required information. \texttt{[...]}\\[1pt]
\textcolor{jbobad}{\ding{55}}~\textit{condition violated} \qquad \textcolor{jbook}{\ding{51}}~\textit{final answer correct}

\end{tcolorbox}
\caption{\textbf{The Worth of A Step Is Settled Downstream.} (a) Before tempering: the output meets its declared condition and gives two images. However, neither carries the information the subtask demands, and the flow reaches the wrong final answer. (b) After tempering: the output that violates the condition confirms that no image qualifies, and the flow reaches the correct final answer. However, if scored against its own condition, the step that helped is recorded as the step that failed.}
\label{fig:appendix:judge_by_outcome}
\end{figure}

%% file: algorithms/global_localization.tex
\begin{algorithm}[h]
\caption{\textsc{Localize}$(W,\mathrm{Mem})$: Global Localization (\S\ref{appendix:subsec:inflowop:global})}
\label{alg:global}
\begin{algorithmic}[1]
\Require workflow $W$ after the local pass of Alg.~\ref{alg:inflowop}, its realized results $\mathrm{Mem}$, global-probe budget $B_{\mathrm{g}}$
\State $\mathcal{B}\gets\varnothing$ \Comment{probed suspects}
\While{no usable final answer \textbf{and} $\lvert\mathcal{B}\rvert<B_{\mathrm{g}}$}
  \State $s\gets\arg\max_{s\notin\mathcal{B}}\mathrm{susp}(s)$;\quad $\mathcal{B}\gets\mathcal{B}\cup\{s\}$ \Comment{Eq.~\ref{eq:suspicion}: global only localizes}
  \State $b\gets\textsc{Detect}(s,\mathrm{Mem}[s])$ \Comment{a suspected but clean subtask is left alone}
  \If{$b\neq\textit{none}$}
     \State $W\gets\textsc{Temper}(W,s,b)$;\quad $\mathrm{Mem}\gets\mathrm{Mem}\setminus I(s)$ \Comment{re-walk the closure: the same local correction}
  \EndIf
\EndWhile
\State \Return final answer;\quad $W^{\dagger}\gets W$
\end{algorithmic}
\end{algorithm}

%% file: figures/exp_cost.tex
\begin{figure}[H]
    \small
    \centering

    \includegraphics[width=\linewidth]{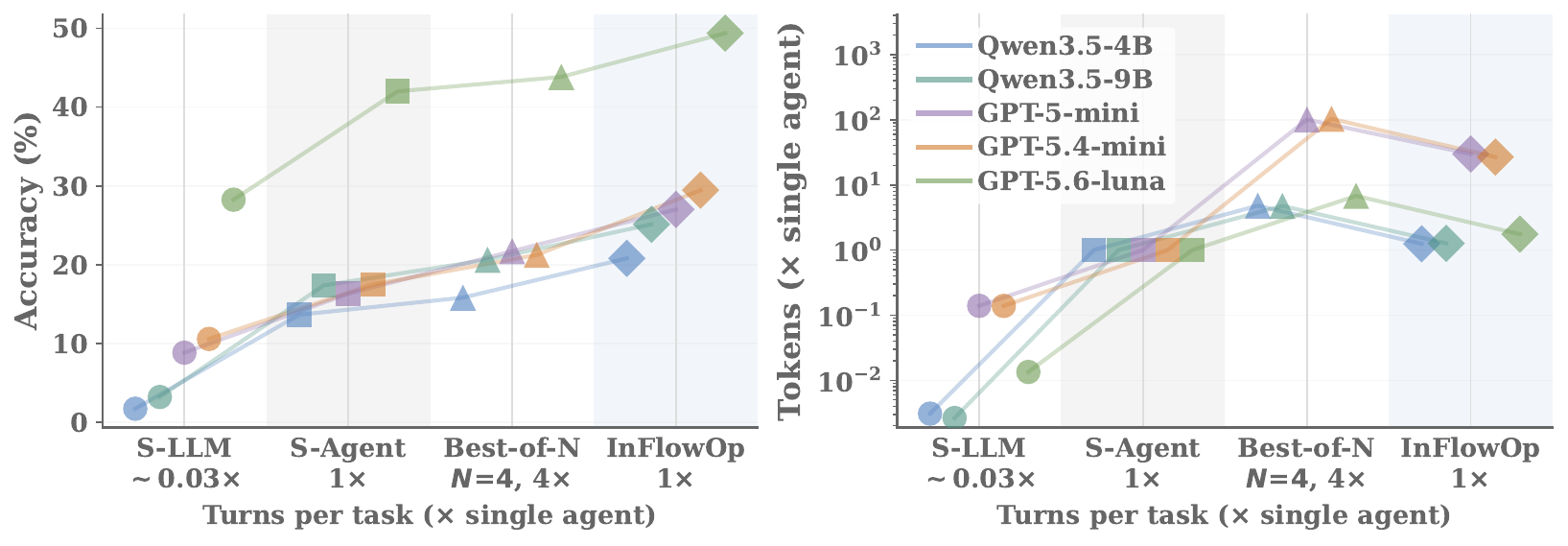}

    \caption{\textbf{What Each Method Spends, and What It Earns.} Mean accuracy (\%) over the $16$ arms of Tab.~\ref{tab:exp_main} (\textit{left}), and the tokens a solver spends per task over the same arms (\textit{right}, logarithmic), with one color per backbone. \textit{S-LLM} and \textit{S-Agent} are the single-LLM and single-agent baselines (\S\ref{subsec:exps:setup}). \textit{Best-of-N} repeats workflow construction and execution $N\!=\!4$ times and keeps its best outcome. Both costs are stated as a multiple of one single-agent run on the same backbone. Each method's turn cost is given beneath its name.}
    \label{fig:exp_cost}

\end{figure}

%% file: tables/appendix_configuration.tex
\begin{table}[t]
\centering
\small
\caption{\textbf{Experiment Configurations.} We summarize our experiment configurations for core parameters (\S\ref{sec:exps}), and an ablation arm is studied against the configuration stated here.}
\label{tab:appendix:configuration}

\vspace{2pt}

\newcolumntype{C}[1]{>{\hsize=#1\hsize\centering\arraybackslash}X}
\newcolumntype{Y}[1]{>{\hsize=#1\hsize\raggedright\arraybackslash}X}
\begin{tabularx}{\textwidth}{@{}C{0.82}Y{1.26}C{0.92}@{}}
\toprule
\textbf{Parameter} & \multicolumn{1}{c}{\textbf{Definition}} & \textbf{Configuration} \\

\midrule

\rowcolor{blue_bg}\multicolumn{3}{c}{\textbf{\textit{\textcolor{blue_text}{Decomposition}}}} \\

\noalign{\vskip 2pt}

Control-flow primitives & The structures a workflow may use beyond plain sequencing (\S\ref{appendix:subsec:inflowop:meta_primitives}) & \textsc{And}, \textsc{Repeat}, \textsc{Select}, \textsc{Or} \\
Reliability--latency weight $\beta$ & What a unit of latency is worth against a unit of reliability in the workflow cost (Eq.~\ref{eq:workflow_cost}) & $0.25$ \\
Creation penalty $\gamma$ & What creating one agent costs the same objective (Eq.~\ref{eq:workflow_cost}) & $1.0$ \\
Competence bar $\bar C$ & The cost an agent clears to count as covering an atom (Eq.~\ref{eq:coverage}) & $0.7$ \\
Agent creation & Whether a new agent may be created where no pooled agent covers a subtask & on \\
Cost matrix & Whether a created agent enters the matrix during \coalesce, or the matrix stays as first scored & dynamic \\
Expansion budget $B$ & How many groupings \coalesce may expand before it returns its incumbent (\S\ref{subsec:method:decomposition}) & $20{,}000$ \\
Repetition budget $B_{\mathrm{rep}}$ & How many times \textsc{Repeat} may re-run a subtask in place (Eq.~\ref{eq:mp_repeat}) & $1$ \\
\noalign{\vskip 2pt}
\rowcolor{blue_bg}\multicolumn{3}{c}{\textbf{\textit{\textcolor{blue_text}{Estimator}}}} \\
\noalign{\vskip 2pt}
Estimator & How well an agent meets an atom's demands is scored (\S\ref{subsec:method:decomposition}) & rubric \\
Scored on & The evidence about an agent supplied to the estimator & card and profile \\
\noalign{\vskip 2pt}
\rowcolor{blue_bg}\multicolumn{3}{c}{\textbf{\textit{\textcolor{blue_text}{In-Flow Optimization}}}} \\
\noalign{\vskip 2pt}
Scope & Where a fault is looked for once execution falls short (\S\ref{subsec:method:inflowop}) & local \\
Unmet bar $\bar u$ & How far a declared condition falls short before it counts as unmet (Eq.~\ref{eq:credit}) & $0.8$ \\
Re-assign budget $B_{\mathrm{ra}}$ & How many alternative agents a faulty subtask may be given before the ladder climbs (Alg.~\ref{alg:inflowop}) & $1$ \\
Re-decompose budget $B_{\mathrm{rd}}$ & How many times a faulty subtask may be re-decomposed before it reverts (Alg.~\ref{alg:inflowop}) & $1$ \\
Global-probe budget $B_{\mathrm{g}}$ & How many suspected subtasks the optional global pass may probe (Alg.~\ref{alg:global}) & $3$ \\
\bottomrule
\end{tabularx}
\end{table}